# RECAST: A Machine-Learning Framework for Correction and Super-Resolution of Coarse-Grid PDE Solvers

Maryam Reza*, Farbod Faraji**

* Independent Researcher

** Department of Computing, Huxley Building, Imperial College London, London SW7 2RH, United Kingdom

**Abstract**: Coarse-grid numerical solvers can substantially reduce the computational cost of time-dependent PDE simulation, but under-resolution often degrades both the trajectory and the spatial fidelity of the solution. We introduce RECAST (Recurrent Error Correction And Super-resolution of coarse-grid Trajectories), a machine-learning framework designed to restore this lost accuracy while retaining coarse-grid evolution. RECAST combines learned correction within the numerical time-stepping loop with reconstruction of the corresponding fine-grid state from the corrected coarse history. We evaluate the framework on six one-dimensional PDE systems spanning transport, diffusion, dispersion, reaction, and wave dynamics, using spatial grids coarsened by factors of 8–16 and 1000-step closed-loop rollouts from unseen initial conditions. Across the test cases, RECAST remains closely aligned with the fine-grid reference solutions and reduces time-averaged relative error by approximately 50–92% compared with the corresponding uncorrected coarse-grid solvers. Additional tests show generalization to unseen PDE parameter values, while comparison with a contemporary coarse-correction architecture shows that RECAST achieves lower error and better long-horizon agreement with the fine-grid reference over 5000-step rollouts. These results demonstrate that the learned correction and reconstruction capabilities of RECAST can enable substantially coarser PDE evolution without the corresponding loss of solution fidelity, providing a proof-of-concept route toward machine-learning acceleration of higher-dimensional numerical simulations across science and engineering. Code: github.com/MariRe1992/recast.

## 1. Introduction

Computationally efficient numerical simulation of partial differential equations (PDEs) is often limited by the need to resolve small spatial scales over long time horizons. Solving on a coarser grid reduces computational cost but introduces two distinct sources of error. The first is representation error: the coarse grid cannot explicitly capture fine-scale spatial structures. The second is evolution error: the coarse solver advances the resolved scales on the basis of under-resolved dynamics, which can produce numerical diffusion, phase and dispersion errors, incorrect wave speeds, and the omission of sub-grid interactions. Recovering fine-grid dynamics from coarse-grid computation thus requires more than post-processing resolution enhancement. It needs both reconstruction of the missing spatial content *and* correction of the coarse trajectory during time integration, before these errors accumulate.

**Machine-learning super-resolution and neural operators**. Modern machine-learning approaches to super-resolution were developed primarily in computer vision, where convolutional networks, sub-pixel upsampling, and adversarial losses were used to map low-resolution images to high-resolution counterparts [1]-[3]. These ideas were subsequently adapted to scientific computation and fluid mechanics, as part of the broader use of machine learning for simulation acceleration and reduced-order modeling across computational physics [4][5]. Fukami et al. showed that convolutional neural networks (CNNs) and hybrid downsampled skip-connection/multiscale architectures can reconstruct high-resolution turbulent fields from very coarse inputs, with the multiscale design aiding recovery of vortical structures across different spatial scales [6]. Their later spatiotemporal extension used coarse information in both space and time to reconstruct high-resolution turbulent-flow evolution, demonstrating that temporal context can improve the reconstruction of dynamically evolving structures compared with single-snapshot mappings [7]. This is an important distinction in that many CNN-based super-resolution models operate on static input–output pairs, whereas the fine-scale state of a time-dependent PDE may be ambiguous from a single coarse snapshot but become inferable from the advection, diffusion, dispersion, and phase propagation recorded over a temporal history.

Generative adversarial networks (GANs), including SRGAN- and ESRGAN-based formulations [2][3], have also been applied to turbulent velocity-field reconstruction and can recover sharper small-scale structures and improved flow statistics compared with purely regression-based models [8]. Physics-informed enhanced SRGAN approaches have further incorporated physical residuals and subfilter-modeling objectives, particularly for turbulent reactive flows [9]. These methods can be powerful *offline* reconstruction tools, but they are normally applied after the low-resolution trajectory has already been generated. Thus, if the coarse solver has drifted from the fine-grid dynamics, the super-resolution model may reconstruct plausible fine-scale structure while remaining dynamically inaccurate because its input is already biased.

A related line of work concerns neural operators. DeepONet learns nonlinear solution operators through branch and trunk networks, while Fourier neural operators (FNOs) parameterize operator kernels in Fourier space [10], [11]. Because these models learn mappings between function spaces, they are designed to operate across discretizations; FNOs have, for example, demonstrated zero-shot super-resolution on turbulent-flow benchmarks [11]. Operator-based methods have also been proposed for arbitrary-scale image super-resolution [12], while multigrid tensorized FNOs address the memory and scalability constraints associated with high-resolution PDE learning [13]. More recent generative and adversarial neural-operator methods seek to mitigate the oversmoothing of fine-scale turbulent structures produced by conventional $L^2$-trained operators [14]. In the present context, however, these formulations generally serve as surrogate solvers, forecasting models, or direct reconstruction maps rather than as corrective components embedded within a prescribed coarse numerical integrator.

**Learned correction of coarse numerical solvers**. Machine learning has also been used to improve the coarse solver itself. Data-driven discretization learns optimized derivative approximations or finite-difference closures directly on coarse grids, enabling accurate integration at resolutions considerably coarser than those supported by standard schemes [15]. In turbulence and multiscale modeling, neural networks have been trained to approximate sub-grid stresses and closure terms from coarse variables, with both a priori and a posteriori strategies considered for large-eddy simulation [16][17]. Machine-learning-accelerated computational fluid dynamics has further shown that neural corrections embedded within conventional solvers can reproduce the accuracy of much finer simulations at substantially reduced cost while retaining the underlying numerical structure [18]. Related hybrid approaches, such as the Corrective Source Term Approach, augment physics-based models with learned source terms that compensate for unresolved or misspecified physics [19].

Solver-in-the-loop training addresses a central weakness of one-step supervised correction. A model trained only on exact reference states encounters a shifted input distribution when it is applied recursively to its own corrected trajectory. Differentiating through the solver and unrolling multiple time steps exposes the learned correction to states representative of deployment and can substantially improve long-term accuracy and stability [20][21]. Graph-network simulators and physics-informed recurrent architectures extend learned time evolution to mesh-based and autoregressive PDE modeling, although they generally act as simulators themselves rather than as corrections to a prescribed coarse numerical integrator [22][23].

A particularly relevant recent method is $P^2C^2$Net, which combines a trainable PDE block with a neural correction block to solve spatiotemporal PDEs on coarse grids while preserving the PDE-based update structure [24]. $P^2C^2$Net demonstrates that physics-encoded coarse correction can substantially improve prediction accuracy in small-data regimes. Its emphasis, however, is on correcting the coarse-grid state; it is not designed to reconstruct the corresponding fine-grid field. Conversely, offline super-resolution can reconstruct fine-scale content from a low-resolution state but cannot prevent an independently evolved coarse solver from accumulating dynamical error.

**RECAST framework and paper's contributions**. The present work addresses this combined problem by coupling two learned components with complementary roles. A super-resolution model learns a nonlinear map from coarse-grid information to fine-grid states, reducing representation error by recovering the sub-grid spatial content. A separate correction model is inserted within the coarse-solver loop, reducing evolution error by modifying the provisional coarse update before it is fed back into the next time step. These components form RECAST – Recurrent Error Correction And Super-resolution of coarse-grid Trajectories – an ML-integrated hybrid solver framework that enables substantial coarsening of the computational grid while maintaining both the dynamical consistency and spatial resolution of the corresponding fine-grid solution. The correction model is first trained on one-step coarse-grid update errors and then refined through solver-in-the-loop rollout training, so that it learns to reduce accumulated trajectory error rather than only instantaneous residual error. The resulting framework hence differs from purely offline super-resolution, direct neural-operator surrogates, and coarse-grid correction alone, which were reviewed above: it simultaneously corrects the low-resolution dynamical trajectory and reconstructs the corresponding fine-resolution PDE state.

This distinction is also crucial for interpreting the results and judging the practical applicability of the method. In conventional super-resolution, the low-resolution input is often obtained by downsampling a high-resolution reference and is therefore dynamically consistent with the target. In the present setting, the low-resolution state is generated recursively by an independently evolved coarse solver, so the input contains both missing spatial information and accumulated dynamical bias, as would occur in an actual coarse-grid simulation.

Both the super-resolution and correction components of RECAST are based on the Shallow Recurrent Decoder (SHRED) architecture [25]. SHRED was originally developed as a sparse-sensing architecture for reconstructing high-dimensional states from limited time-history measurements. Its key idea is to replace static snapshot-based sensing with temporal sensor trajectories, allowing the recurrent encoder to extract dynamical information from a few measurements and a shallow decoder to reconstruct the full high-dimensional state from the encoder's latent space [25]. In its original demonstrations, SHRED reconstructed turbulent-flow fields from very limited, often randomly placed sensors and outperformed POD/gappy-POD and shallow-decoder baselines while showing reduced dependence on optimized sensor placement [25]. Subsequent work extended SHRED to plasma systems, including the reconstruction of full high-dimensional plasma states from minimal local or global measurements, the inference of multiple coupled plasma fields from measurements of only one accessible quantity, and reduced-order modeling of complex low-temperature $E \times B$ plasma dynamics [26][27][28]. Recent theoretical work has further established conditions and bounds under which temporal histories can substitute for spatial measurements while retaining a prescribed level of information about the underlying dynamical state [29]. SHRED's success in compressed-sensing applications and its ability to reconstruct high-dimensional states from sparse time-history measurements thus motivate its use here as the shared network backbone within RECAST: SHRED-SR performs coarse-to-fine solution reconstruction, and SHRED-delta performs solver-integrated correction, with the coarse solver acting as the "sparse" measurement input.

RECAST is evaluated in this paper across six one-dimensional PDE systems at spatial coarsening factors of $8\times$ and $16\times$ over 1000-step closed-loop rollouts. It reduces the time-averaged relative error by approximately 50-92% compared with the uncorrected coarse solver. Additional tests assess parameter-dependent generalization and long-horizon behavior, including a 5000-step comparison with an adapted $P^2C^2$Net baseline. The results support the central premise of RECAST: when the low-resolution input is produced by an independently evolved numerical solver, controlling the accumulated evolution error is essential for reliable fine-grid reconstruction.

## 2. Methodology

Building on the framework introduced in Section 1, this section specifies the construction of the paired coarse–fine data, the SHRED architecture used by both learned components, their task-specific training procedures, and the post-training deployment of RECAST.

RECAST uses two recurrent SHRED networks [25], described in Section 2.2. SHRED-SR receives a temporal history of coarse-grid states and returns the corresponding fine-grid state, as formulated in Section 2.3. SHRED-delta receives a coarse-state history ending with the provisional update produced by the numerical solver and returns the correction applied before the corrected state is fed back into the next time step, as formulated in Section 2.4.

The workflow first generates high-resolution reference trajectories and projects them onto a coarse grid to construct paired fine- and coarse-grid training data. Training then proceeds in three stages. First, SHRED-SR is trained to learn the mapping from coarse-grid histories to fine-grid states. Next, SHRED-delta is pretrained using one-step supervised learning, where the target is the difference between the next fine-grid state projected onto the coarse grid and the provisional state predicted by the coarse solver. Finally, SHRED-delta is refined in a solver-in-the-loop (SOL) training stage, in which its corrections are recursively inserted into the coarse-grid solver over multiple time steps [20][21]. The two SHRED-delta training stages hence penalize both instantaneous correction error and accumulated trajectory error relative to the projected fine-grid reference.

During post-training deployment, the method advances the coarse state using the numerical solver, applies the learned SHRED-delta correction, feeds the corrected state into the next time step, and uses SHRED-SR to reconstruct the corresponding fine-grid solution. This produces a dynamically corrected coarse trajectory together with a fine-grid reconstruction, allowing the framework to address both the loss of spatial resolution and the temporal drift caused by coarse numerical evolution.

### 2.1. Mathematical formulation and construction of paired coarse–fine data

We consider a time-dependent PDE defined on a physical domain $\Omega$, written abstractly as

$$\frac{\partial u}{\partial t} = \mathcal{F}(u, x, t; \mu), \quad \text{(Eq. 1)}$$

where $u(x, t; \mu)$ is the solution field and $\mu$ collects the physical parameters as well as the forcing, boundary, and initial conditions, where applicable. We assume that the same physical system is advanced on two spatial discretizations of $\Omega$: a fine grid, denoted by $\Omega_f$, and a coarse grid, denoted by $\Omega_c$. The corresponding discrete solution states are written as

$$u_f^k \in \mathbb{R}^{N_f}, \qquad u_c^k \in \mathbb{R}^{N_c}, \tag{Eq. 2}$$

where $N_f > N_c$, and $k$ is the discrete time index. The fine-grid solver is treated as the reference evolution operator,

$$u_f^{k+1} = \mathcal{R}_f(u_f^k), \tag{Eq. 3}$$

while the coarse-grid solver is written as

$$u_c^{k+1} = \mathcal{S}_c(u_c^k). \tag{Eq. 4}$$

In Eqs. 3 and 4, $\mathcal{R}_f$ denotes the high-fidelity time-advancement rule, and $\mathcal{S}_c$ denotes the low-resolution solver. The coarse solver is computationally efficient, but it leads to loss of sub-grid structure. Importantly, advancing the PDE on an under-resolved grid changes more than the spatial representation of the solution, also altering the time evolution, introducing accumulated numerical diffusion, phase drift, dispersion error, and inaccuracies in the evolution of larger-scale structures due to missing interactions with sub-grid details.

A restriction operator $P: \mathbb{R}^{N_f} \to \mathbb{R}^{N_c}$ maps the fine-grid state to the coarse-grid state space:

$$\bar{u}_c^k = P(u_f^k). \tag{Eq. 5}$$

In the present implementation, $P$ is a block-averaging restriction operator that averages the fine-grid values within each coarse-grid cell; the formulation is nonetheless general and can also accommodate other conservative or physically motivated restriction operators. The paired training data are thus constructed as

$$\{(\bar{u}_c^k, u_f^k)\}_{k=0}^{K}, \tag{Eq. 6}$$

where $\bar{u}_c^k$ is the fine-grid reference projected onto the coarse grid, and $u_f^k$ is the corresponding fine-grid target. The overbar distinguishes the projected reference from the independently evolved coarse-grid state $u_c^k$. The paired trajectories in Eq. 6 are used to construct the lagged input–target sequences for SHRED-SR, as described in Section 2.3.

### 2.2. SHRED architecture as the backbone for the learned components

Both learned components of RECAST use SHRED as the architectural backbone, but they are trained separately for two different tasks: super-resolution reconstruction, which maps a temporal history of coarse-grid states to the corresponding fine-grid state, and solver-integrated correction, which predicts the correction applied to a provisional coarse-grid solver update. Thus, the two networks share the same architectural principle but have different inputs, outputs, parameters, and training objectives.

Let

$$Z^k = (z^{k-q+1}, z^{k-q+2}, \dots, z^k), \tag{Eq. 7}$$

denote a generic lagged input sequence (temporal history) of length $q$. $z^k$ denotes the input state at time $k$. Depending on the task, $Z^k$ represents either a history of coarse-grid states or a coarse-state history whose final element is a provisional update produced by the coarse numerical solver. The sequence $Z^k$ is first passed through a recurrent encoder implemented as a long short-term memory (LSTM) network:

$$h^k = \mathcal{E}_\psi(Z^k), \tag{Eq. 8}$$

where $h^k$ is the latent representation extracted from the temporal history, and $\psi$ denotes the parameters of the generic SHRED mapping. The recurrent encoder processes the input sequence in chronological order to capture the recent dynamical evolution of the coarse field, including propagation, phase information, growth or decay, and other temporal signatures relevant to the reconstruction or correction task.

The latent state is then passed through a shallow nonlinear decoder:

$$\hat{y}^k = \mathcal{Q}_\psi(h^k), \tag{Eq. 9}$$

so that the complete learned mapping can be written as

$$\hat{y}^k = \mathcal{M}_\psi(Z^k) = \mathcal{Q}_\psi\big[\mathcal{E}_\psi(Z^k)\big]. \quad \text{(Eq. 10)}$$

The decoder consists of fully connected nonlinear layers that map the recurrent latent representation to the required output space. The dimension of $\hat{y}^k$ is determined by the task: for the super-resolution model, the output is the fine-grid solution field; for the correction model, the output is a coarse-grid correction vector. In both cases, the recurrent encoder supplies the temporal information, while the decoder learns the nonlinear mapping from the encoded dynamics to the target spatial field or correction. The specific layer sizes and hyperparameters used in the experiments are given in Section 2.6.

This shared architecture is well suited to the present solver-integrated super-resolution problem because the instantaneous coarse-grid state may be insufficient to determine the unresolved fine-scale solution or the required coarse-grid correction. The recent temporal history provides additional dynamical constraints, allowing the model to infer information that is not explicitly available from a single coarse snapshot. Sections 2.3 and 2.4 describe the two task-specific uses of the backbone architecture.

### 2.3. Super-resolution reconstruction model and supervised training

The first learned component of RECAST is the super-resolution reconstruction model, referred to as SHRED-SR and denoted by $\mathcal{G}_\theta$. The purpose of this model is to recover fine spatial structure from the recent coarse-grid dynamics. Using the paired trajectories constructed in Section 2.1, SHRED-SR is trained independently of the coarse-solver time-marching loop to learn the mapping from temporal histories of projected coarse-grid states to the corresponding fine-grid states.

For a history length $q$, the input sequence ending at time $k$ is

$$X^k = \big(\bar{u}_c^{k-q+1}, \bar{u}_c^{k-q+2}, \dots, \bar{u}_c^k\big), \quad \text{(Eq. 11)}$$

The corresponding target is the fine-grid state at the final time of the sequence:

$$Y^k = u_f^k. \quad \text{(Eq. 12)}$$

Accordingly, SHRED-SR represents the nonlinear operator $\mathcal{G}_\theta: (\mathbb{R}^{N_c})^q \to \mathbb{R}^{N_f}$, with prediction:

$$\hat{u}_f^k = \mathcal{G}_\theta(X^k) \approx u_f^k. \quad \text{(Eq. 13)}$$

The model parameters are determined by minimizing a supervised field-reconstruction loss over the training set $\mathcal{D}_{train}$ of lagged input–target pairs:

$$\theta^* = \arg\min_\theta \sum_{(X^k, Y^k) \in \mathcal{D}_{train}} \|Y^k - \mathcal{G}_\theta(X^k)\|_2^2. \quad \text{(Eq. 14)}$$

The data normalization, architecture settings, and optimization parameters used for this training stage are specified in Section 2.6.

### 2.4. Solver-integrated correction and training process

The second learned component of RECAST is the correction model, referred to as SHRED-delta and denoted by $\mathcal{D}_\phi$. It is trained to act inside the coarse-solver loop. At each time step, it predicts the missing coarse-grid correction to be added to the provisional update produced by the coarse-grid solver, so that the corrected state can be fed into the next time step.

Suppose that the current coarse state at time $k$ is $u_c^k$. The coarse-grid solver first produces the provisional update

$$u_{\text{pde}}^{k+1} = \mathcal{S}_c(u_c^k). \quad \text{(Eq. 15)}$$

During training, the corresponding reference state is obtained by projecting the fine-grid state onto the coarse-grid state space using the restriction operator $P$ defined in Section 2.1,

$$\bar{u}_c^{k+1} = P\big(u_f^{k+1}\big). \quad \text{(Eq. 16)}$$

The required coarse-grid correction is then

$$\delta^{k+1} = \bar{u}_c^{k+1} - u_{\text{pde}}^{k+1}. \quad \text{(Eq. 17)}$$

The correction model is trained to approximate $\delta^{k+1}$. Its input is a coarse-state history of length $q$ whose final element is the provisional update:

$$\widetilde{H}_c^{k+1} = \left(u_c^{k-q+2}, \dots, u_c^k, u_{\text{pde}}^{k+1}\right). \quad \text{(Eq. 18)}$$

The predicted correction is

$$\hat{\delta}^{k+1} = \mathcal{D}_\phi\left(\widetilde{H}_c^{k+1}\right), \quad \text{(Eq. 19)}$$

and the corrected coarse state is

$$u_c^{k+1} = u_{\text{pde}}^{k+1} + \hat{\delta}^{k+1}. \quad \text{(Eq. 20)}$$

This corrected state is then inserted into the history buffer and used as the input to the next coarse-grid solver step.

The training process for SHRED-delta consists of two stages, during which SHRED-SR is kept fixed. In the first stage, SHRED-delta is pretrained using one-step correction targets. For each lagged training sequence of projected reference states, the final state $\bar{u}_c^k$ is advanced once using the coarse-grid solver. The fine-grid state at $k+1$ is then projected onto the coarse grid to provide the corresponding reference state. The one-step target defined in Eq. 17 can therefore be written as

$$\delta^{k+1} = \bar{u}_c^{k+1} - u_{\text{pde}}^{k+1} = P\left(u_f^{k+1}\right) - \mathcal{S}_c\left(\bar{u}_c^k\right). \quad \text{(Eq. 21)}$$

The corresponding pretraining objective is

$$\phi_{\text{pre}}^* = \arg\min_\phi \sum_k \left\|\delta^{k+1} - \mathcal{D}_\phi\left(\widetilde{H}_c^{k+1}\right)\right\|_2^2. \quad \text{(Eq. 22)}$$

This stage teaches SHRED-delta the local one-step error introduced by the coarse-grid solver and provides the initialization for solver-in-the-loop refinement.

In the second stage, SHRED-delta is fine-tuned through solver-in-the-loop (SOL) rollout training [20][21]. The correction model is integrated into the coarse-grid solver, and the corrected state is recursively advanced over a rollout horizon of $H_r$ time steps.

For each rollout step,

$$\begin{aligned} u_{\text{pde}}^{k+j+1} &= \mathcal{S}_c\left(u_c^{k+j}\right), \\ \hat{\delta}^{k+j+1} &= \mathcal{D}_\phi\left(u_c^{k+j-q+2}, \dots, u_c^{k+j}, u_{\text{pde}}^{k+j+1}\right), \\ u_c^{k+j+1} &= u_{\text{pde}}^{k+j+1} + \hat{\delta}^{k+j+1}, \qquad j = 0, \dots, H_r - 1. \end{aligned} \quad \text{(Eq. 23)}$$

This training stage exposes SHRED-delta to the same recursive use encountered during deployment and trains it against both instantaneous correction error and accumulated drift of the corrected coarse trajectory.

At each rollout step, the corrected coarse state $u_c^{k+j+1}$ is compared with the corresponding projected reference $\bar{u}_c^{k+j+1} = P\left(u_f^{k+j+1}\right)$. The SOL objective thus penalizes the deviation of the recursively corrected coarse trajectory from the projected fine-grid reference:

$$\phi^* = \arg\min_\phi \sum_k \sum_{j=0}^{H_r-1} \left\|u_c^{k+j+1} - \bar{u}_c^{k+j+1}\right\|_2^2. \quad \text{(Eq. 24)}$$

The SOL training is performed progressively over increasing rollout horizons. Shorter horizons first constrain the correction model to remain stable over a small number of recurrent solver steps, while longer horizons subsequently train it against accumulated multi-step errors. Checkpoint selection is based on the rollout validation error evaluated over a fixed horizon of length $R$:

$$\varepsilon_{\text{rollout}}(R) = \left[ \frac{\sum_{j=1}^{R} \left\| u_c^{k+j} - \bar{u}_c^{k+j} \right\|_2^2}{\sum_{j=1}^{R} \left\| \bar{u}_c^{k+j} \right\|_2^2} \right]^{1/2}. \qquad \text{(Eq. 25)}$$

The rollout horizons and fixed validation horizon used in the experiments are specified in Section 2.6.

Using the same validation rollout length for all SOL training stages allows checkpoints trained with different rollout horizons to be compared on a common recurrent prediction task. The selected SHRED-delta model is the checkpoint with the lowest fixed-horizon rollout validation error. Figure 1 summarizes the complete RECAST training workflow, including SHRED-SR and the one-step pretraining and SOL refinement of SHRED-delta.

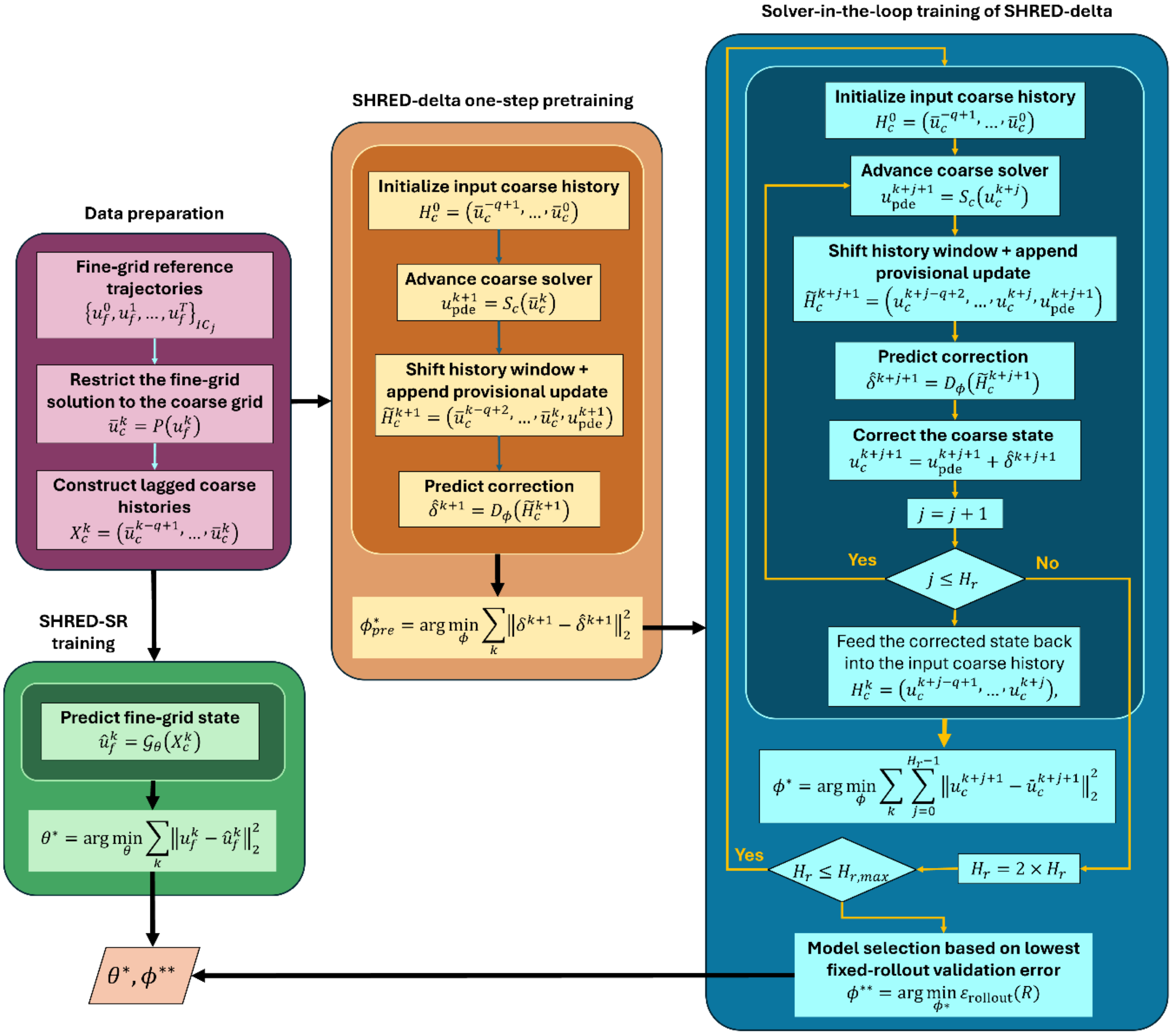

*Figure 1: Training workflow of RECAST. Fine-grid reference trajectories are restricted to the coarse grid to construct lagged history inputs for training. SHRED-SR is trained to reconstruct fine-grid states from projected coarse-grid histories, while SHRED-delta is first pretrained on one-step correction targets and then refined through solver-in-the-loop rollout training. The final SHRED-delta model is selected based on the lowest fixed-horizon rollout validation error.*

### 2.5. Post-training deployment of RECAST

After training, the fixed SHRED-SR and SHRED-delta models are coupled with the coarse-grid solver, and RECAST is advanced recursively on unseen trajectories of the PDE for which the models were trained.

Given an initial coarse-state history $H_c^0$ of length $q$, initialized as described in Section 2.6, each deployment (online) step performs the following update:

$$\begin{aligned} u_{\text{pde}}^{k+1} &= \mathcal{S}_c(u_c^k), \\ \hat{\delta}^{k+1} &= \mathcal{D}_{\phi^*}\left(u_c^{k-q+2}, \ldots, u_c^k, u_{\text{pde}}^{k+1}\right), \\ u_c^{k+1} &= u_{\text{pde}}^{k+1} + \hat{\delta}^{k+1}, \\ \hat{u}_f^{k+1} &= \mathcal{G}_{\theta^*}\left(u_c^{k-q+2}, \ldots, u_c^{k+1}\right). \end{aligned} \qquad \text{(Eq. 26)}$$

The corrected coarse state $u_c^{k+1}$ is fed back into the next coarse-grid solver step, and $\hat{u}_f^{k+1}$ is the corresponding reconstructed fine-grid output. Figure 2 presents the closed-loop deployment of RECAST.

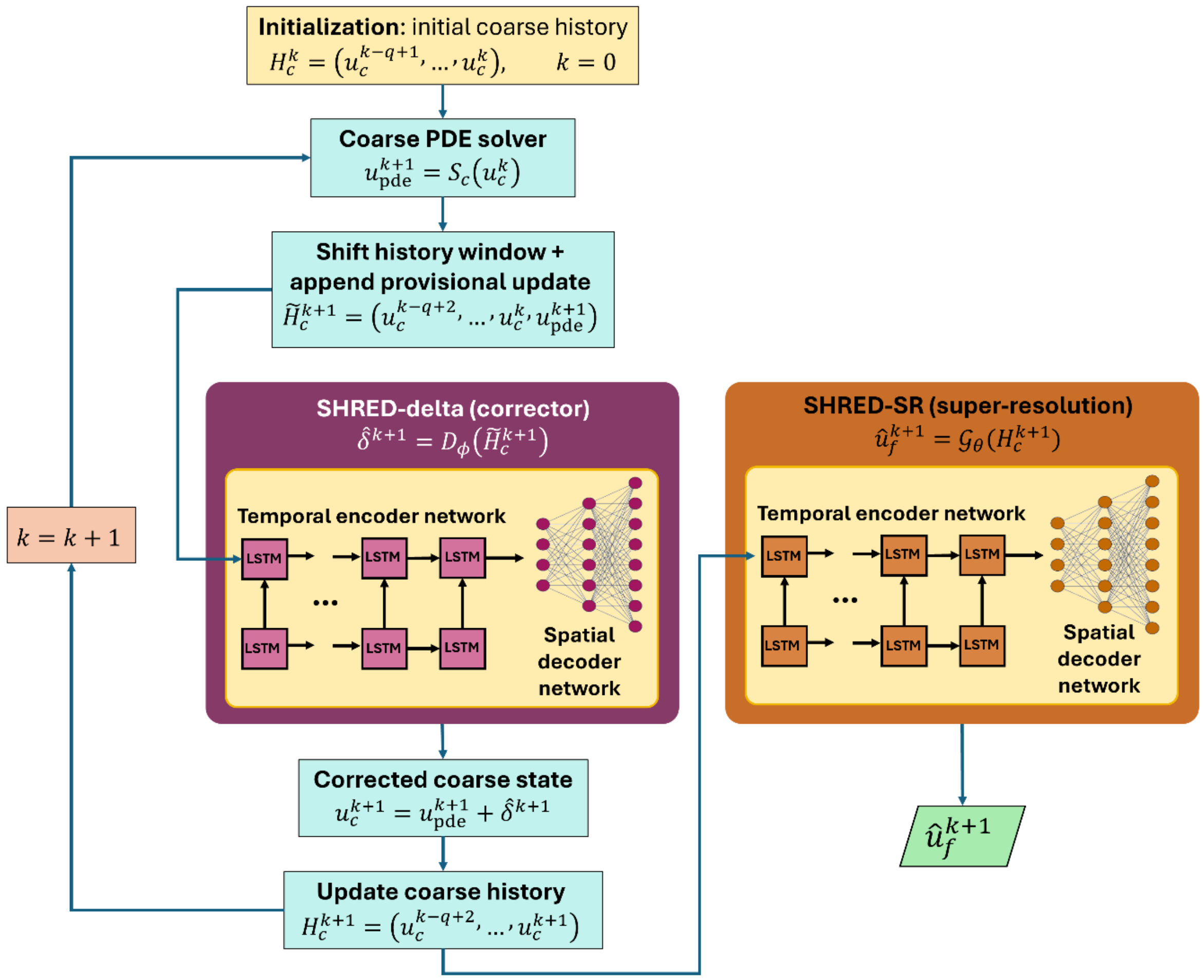

*Figure 2: Closed-loop deployment of RECAST. At each time step, the coarse-grid PDE solver produces a provisional update, SHRED-delta predicts the corresponding correction, and the corrected coarse state is fed back into the next solver step. SHRED-SR then maps the updated corrected coarse-state history to the reconstructed fine-grid state.*

### 2.6. Model architecture, data, and training setup

All numerical experiments follow the same data-processing, backbone hidden-architecture, training, and validation protocol, while the PDE solver, physical parameters, and coarse-grid dimension vary by case. For each PDE, the reference data are generated as an ensemble of independent fine-grid trajectories with different initial conditions and, where applicable, physical parameter values. In the present setup, each dataset contains 500 trajectories with 1000 recorded time instances per trajectory, resulting in $5 \times 10^5$ snapshots. The fine grid contains 256 spatial points, while the coarse grid contains 32 or 16 spatial points, depending on the PDE case. The corresponding full fine- and coarse-grid state dimensions are denoted by $N_f$ and $N_c$, respectively. The corresponding projected coarse-grid references are obtained using the block-averaging restriction $P$ defined in Section 2.1, which averages the fine-grid values within each non-overlapping coarse-grid cell.

The temporal input to both SHRED-SR and SHRED-delta is a lagged coarse-state sequence of length $q = 50$. Valid lagged sequences are constructed separately within each independent trajectory, so no temporal input crosses from one trajectory to another. Using the indexing convention adopted in the dataset construction, 950 valid lagged samples are retained per trajectory. The trajectories, rather than individual time steps or lagged samples, are split into training, validation, and test sets using proportions of 70%, 15%, and 15%, respectively. This gives 350 training trajectories, 75 validation trajectories, and 75 test trajectories, corresponding to 332,500 training samples, 71,250 validation samples, and 71,250 test samples for supervised SHRED-SR training. The same trajectory-level partition is used throughout the subsequent training stages.

Before training, separate feature-wise min-max scalers are fitted to the coarse-grid inputs, fine-grid targets, and correction targets using only the training trajectories,

$$\tilde{x} = \frac{x - x_{\min}}{x_{\max} - x_{\min}} \quad \text{(Eq. 27)}$$

where $x_{\min}$ and $x_{\max}$ are obtained from the corresponding training data. Scaling is applied independently to each feature of the relevant flattened state or correction vector. The fitted scaling parameters are then fixed and applied unchanged to the validation and test trajectories. During deployment, the coarse-grid solver advances the state in physical units. Each coarse-state history is normalized before network evaluation, while the predicted correction and fine-grid reconstruction are inverse-transformed to physical units before the correction is applied or the reconstruction is reported.

Both learned models use the same SHRED backbone structure described in Section 2.2, with task-specific output dimensions. The recurrent encoder is a two-layer LSTM with hidden size 64. The decoder contains two fully connected hidden layers of widths 350 and 400, with ReLU activation after each hidden layer and a dropout probability of 0.1 during training. The SHRED-SR output dimension is $N_f$, corresponding to the full fine-grid state, while the SHRED-delta output dimension is $N_c$, corresponding to the coarse-grid correction. Both models receive inputs of shape $q \times N_c$.

SHRED-SR is trained first through supervised reconstruction of the fine-grid states from lagged coarse-grid histories. Training uses the Adam optimizer [30] with starting learning rate $10^{-3}$, weight decay of $10^{-5}$, batch size of 256, and gradient clipping with maximum norm of 1.0. Training is performed for at most 50 epochs, with validation evaluated every two epochs and early stopping after 10 consecutive validation evaluations without improvement. Validation is based on the relative error between the predicted and true scaled fine-grid fields. A validation-based adaptive learning-rate schedule reduces the learning rate by a factor of 0.5 when the validation error stops improving. The model parameters giving the lowest validation error are retained.

SHRED-delta is first pretrained for 50 epochs on supervised one-step correction targets using the same Adam hyperparameters, batch size, and gradient-clipping threshold as SHRED-SR. The pretrained parameters giving the lowest validation loss are used to initialize the SOL training stage. SOL training follows a staged rollout curriculum with horizons $H_r = 2$, $H_r = 4$, and $H_r = 8$, using 400 optimizer iterations at each horizon. The maximum horizon is set to $H_r = 8$, since longer horizons were not observed to improve training. Model selection is based on the fixed-length rollout error defined in Eq. 25 in Section 2.4 rather than on the training objective at the current curriculum horizon. The fixed validation rollout length is $R = 100$ time steps and is kept unchanged across all checkpoints and SOL stages. This allows checkpoints trained at different rollout horizons to be compared on the same recurrent prediction task. The selected SHRED-delta model is the checkpoint with the lowest fixed-rollout relative trajectory error on the validation trajectories.

After training, evaluation is performed in two complementary settings. First, the standalone SHRED-SR model is evaluated offline on held-out test samples. Second, RECAST is evaluated in closed-loop on unseen trajectories. A reference coarse-grid history of length $q = 50$ initializes each rollout; these initialization states are not included in the reported rollout error. The coarse solver, SHRED-delta correction, and SHRED-SR reconstruction are then applied recursively using unseen initial conditions and, in the parameter-generalization experiment, unseen PDE parameter values. Performance is assessed against both the uncorrected coarse-grid baseline and the corresponding fine-grid reference over the same time interval.

## 3. Numerical experiments and results

The method is demonstrated on six one-dimensional PDE test cases defined in Table 1 and Table 2, selected to span several ways in which coarse-grid discretization can distort PDE evolution and physically important solution features.

The variable-coefficient advection–diffusion equation represents transport in a heterogeneous medium; coarse grids can misrepresent both spatially varying propagation speeds and diffusive behavior. The Korteweg–de Vries-Burgers equation represents nonlinear dispersive–diffusive wave motion, creating competing steepening, dispersion, and viscous damping, where under-resolution produces phase errors, incorrect pulse damping, and distorted dispersive tails. The wave-propagation problem in inhomogeneous dielectric media represents transport through spatially varying material properties, where coarse grids can introduce wave-speed and interface-delay errors and excessive numerical smoothing across the dielectric slab. The FitzHugh–Nagumo system represents excitable reaction–diffusion dynamics, where coarse grids can smear traveling activation fronts and distort the coupled activator–recovery structure. The linear Schrödinger equation models complex-valued dispersive evolution, interference, and reflection from a smooth potential, where coarse resolution can alter phase accumulation and interference patterns. Finally, the shallow-water/Saint-Venant system represents nonlinear height–momentum dynamics over bottom topography, where coarse grids can distort wave propagation, reflection, and topography-induced wave trains.

These cases span linear and nonlinear dynamics, heterogeneous coefficients, dispersion, diffusion, reaction kinetics, wave propagation, and balance-law effects, providing a broad test set for evaluating RECAST across distinct sources of coarse-grid error.

| PDE | Equation | Parameters |
|---|---|---|
| **Advection–diffusion** | $u_t + \partial_x(a(x)u) = \partial_x(\kappa(x)u_x)$ | $a(x) = \max\left\{c_{\mathrm{adv}}\left[1 + 0.5\sin\left(\frac{2\pi x}{L} + 0.25\right) + 0.5\sin\left(\frac{4\pi x}{L} - 0.70\right)\right], 0.2c_{\mathrm{adv}} + 10^{-6}\right\}, c_{adv} = 1.0$<br>$\kappa(x) = \max\left\{\nu\left[1 + \sum_{m=1}^{3} A_m\, e^{-\frac{1}{2}(d_L(x,x_m)/\sigma_m)^2}\right], 10^{-8}\right\}, \nu = 0.002;$<br>$(x_m/L, \sigma_m/L, A_m) = (0.18, 0.055, 1.5), (0.52, 0.075, 3.0), (0.78, 0.045, 4.5).$ |
| **KdV–Burgers** | $u_t + c_{adv}uu_x + u_{xxx} = \nu u_{xx}$ | $c_{adv} = 6.0,\ \ \nu = 0.1.$ |
| **Electric-field propagation in an inhomogeneous dielectric** | $E_t = -\partial_x(c(x)E) + \nu E_{xx},$<br>$c(x) = 1/\sqrt{\epsilon(x)}$ | $\nu = 2\times 10^{-4};$<br>$\epsilon(x) = 1 + \epsilon_r s(x),\ s(x) = \frac{1}{2}\left[tanh\left(\frac{x-0.45L}{0.015L}\right) - tanh\left(\frac{x-0.68L}{0.015L}\right)\right], \epsilon_r = 10.0.$ |
| **FitzHugh–Nagumo reaction-diffusion** | $u_t = D_u u_{xx} + u - u^3/3 - v;$<br>$v_t = D_v v_{xx} + \varepsilon(u + a - bv)$ | $D_u = 1.0,\ D_v = 0.05;$<br>$\varepsilon = 0.08,\ a = 0.70,\ b = 0.80.$ |
| **Schrödinger** | $i\psi_t = -D\psi_{xx} + V(x)\psi$ | $D = 1.0;$<br>$V(x) = V_0\left[\exp\left(-\frac{1}{2}\left(\frac{x+0.32L}{0.035L}\right)^2\right) + \exp\left(-\frac{1}{2}\left(\frac{x-0.32L}{0.035L}\right)^2\right) + 0.08\exp\left(-\frac{1}{2}\left(\frac{x}{0.18L}\right)^2\right)\right],\ V_0 = 8.0.$ |
| **Shallow-water / Saint-Venant** | $h_t + q_x = \nu_h h_{xx};$<br>$q_t + \partial_x\left(\frac{q^2}{h} + 0.5gh^2\right) = -ghb_x - rq + \nu_q q_{xx}$ | $\nu_h = 0.01,\ \nu_q = 0.02,\ g = 1.0, r = 0.015, A_b = 0.18;$<br>$b(x) = \tilde{b}(x) - \min_x \tilde{b}(x),$<br>$\tilde{b}(x) = A_b\left[\exp\left(-\frac{1}{2}\left(\frac{x+0.18L}{0.060L}\right)^2\right) + 0.75\exp\left(-\frac{1}{2}\left(\frac{x-0.22L}{0.075L}\right)^2\right) + 0.18cos\left(\frac{2\pi x}{L} + 0.4\right) + 0.08cos\left(\frac{4\pi x}{L} - 0.8\right)\right].$ |

*Table 1: Governing equations and physical parameters for the PDE test cases used to evaluate RECAST.*

For each PDE, reference trajectories are generated from independently sampled initial conditions. Unless otherwise stated, the parameters defining these initial conditions are sampled independently from the ranges or discrete sets specified in Table 2. Here,

$$G(x; x_j, \sigma_j) = \exp\left[-\frac{1}{2}\left(\frac{d_L(x, x_j)}{\sigma_j}\right)^2\right], \qquad \text{(Eq. 28)}$$

denotes a periodic Gaussian pulse centered at $x_j$ with width $\sigma_j$, where $d_L$ is the shortest periodic distance on a domain of length $L$. The symbols $\mathcal{N}_\infty$, $\mathcal{N}_0$, and $\mathcal{N}_{rms}$ denote normalization operators, defined as

$$\mathcal{N}_\infty[f] = \frac{f}{\max_x |f|}, \quad \mathcal{N}_0[f] = \frac{f - \langle f\rangle}{\max_x |f - \langle f\rangle|}, \quad and \quad \mathcal{N}_{rms}[f] = \frac{f}{\sqrt{\langle |f|^2\rangle}}, \qquad \text{(Eq. 29)}$$

where $\langle\cdot\rangle$ denotes the spatial average. All test cases use periodic boundary conditions. For each case, the model is tested on trajectories generated from unseen initial conditions withheld from training. The central question is whether a fine-grid solution can be reconstructed from a coarse representation and whether the coarse numerical trajectory can be corrected sufficiently for recursive time integration to remain dynamically close to the fine-grid reference. This is an important distinction to mind because purely post-processing – referred to here as offline super-resolution – can improve the visual fidelity of individual snapshots while still allowing the underlying coarse solver to drift from the fine-grid reference trajectory.

| PDE | Numerical method | Initial conditions | L | Fine-grid points | Coarse-grid points | Δt |
|---|---|---|---|---|---|---|
| **Advection–diffusion** | Conservative finite-volume RK4. | $u_0 = \mathcal{N}_\infty\left[\sum_{j=1}^{4} A_j\, G(x; x_j, \sigma_j) + 0.25\sin(2\pi x/L + \phi_1) + 0.15\sin(4\pi x/L + \phi_2) + 0.08\cos(6\pi x/L + \phi_3)\right];$ with $x_j \in [0, L]$, $\sigma_j \in [0.025L, 0.06L]$, $A_j \in [-1,1]$, and $\phi_1, \phi_2, \phi_3 \in [0,2\pi]$. | 2π | 256 | 32 | 0.01 |
| **KdV–Burgers** | Fourier pseudo-spectral ETDRK4, 2/3 de-aliasing. | $u_0 = \mathcal{N}_0\left[\sum_{j=1}^{2} A_j\, G(x; x_j, \sigma_j) + 0.2\sin(2\pi x/L + \phi_1) + 0.1\sin(4\pi x/L + \phi_2)\right],$ with $x_j \in [0, L]$, $\sigma_j \in [0.65,1.35]$, $A_j = s_j\alpha_j$, where $\alpha_j \in [0.5,1.2]$, and $s_j \in \{-1,1\}$ with equal probability, and $\phi_1, \phi_2 \in [0,2\pi]$. | 20 | 256 | 16 | 0.005 |
| **Electric-field propagation in an inhomogeneous dielectric** | Fine: RK4 spectral derivatives. Coarse: upwind finite volume + central diffusion. | $E_0 = \mathcal{N}_0[G(x; x_0, \sigma)cos(2\pi m(x - x_0)/L + \phi) + 0.05sin(2\pi x/L + \theta)],$ with $x_0 \in [0.05L, 0.3L]$, $\sigma \in [0.05L, 0.3L]$, $m \in \{5,6,7,8,9\}$, $\phi, \theta \in [0,2\pi]$. | 1 | 256 | 32 | 0.001 |
| **FitzHugh–Nagumo reaction-diffusion** | RK4 + central finite differences. | $u_0 = u_* + \sum_{j=1}^{2} A_j^u\, G(x; x_j, \sigma_j) + 0.03\cos(2\pi x/L + \phi),$ $v_0 = v_* + \sum_{j=1}^{2} A_j^v\, G(x; x_j, \sigma_j) + 0.01\sin(2\pi x/L + \phi),$ with $u_* \approx -1.199$, $v_* \approx -0.624$, $x_1 \in [-0.35L, -0.12L]$, $x_2 \in [0.05L, 0.32L]$, $\sigma_j \in [0.04L, 0.065L]$, $A_j^u \in [1.8,2.4]$, $A_j^v \in [0,0.1]$, $\phi \in [0,2\pi]$. | 100 | 256 | 32 | 0.02 |
| **Schrödinger** | Fourier split-step. | $\psi_0 = 0.45\mathcal{N}_{\text{rms}}[P_1 + 0.25P_2]$, where $P_j = G(x; x_j, \sigma_j)\exp[i(q_j(x - x_j) + \phi_j)]$. Main packet: $x_1 \in [-0.18L, -0.04L]$, $\sigma_1 \in [0.050L, 0.075L]$, $q_1 \in [1.2,1.9]$. Broad packet: $x_2 \in [0.02L, 0.14L]$, $\sigma_2 \in [0.08L, 0.12L]$, $q_2 \in [-0.35,0.35]$. $\phi_1, \phi_2 \in [0,2\pi]$. | 40 | 256 | 32 | 0.05 |
| **Shallow-water / Saint-Venant** | Rusanov finite volume + SSPRK3. | Initial free−surface level: $\eta_0 = 1.15 + \max b + A_1 G(x; x_1, \sigma_1) + A_2 G(x; x_2, \sigma_2)$, $h_0 = \max(\eta_0 - b(x), 0.08)$, $q_0 = 0.012\sin(2\pi x/L + \phi)$. With $x_1 \in [-0.38L, -0.18L]$, $\sigma_1 \in [0.055L, 0.09L]$, $A_1 \in [0.20,0.34]$; $x_2 \in [0.02L, 0.28L]$, $\sigma_2 \in [0.06L, 0.11L]$, $A_2 \in [-0.08,0.14]$, $\phi \in [0,2\pi]$. | 30 | 256 | 16 | 0.05 |

*Table 2: Numerical setup for the PDE test cases, including numerical solver method, initial-condition distribution, domain length, fine-grid and coarse-grid sizes, and time step.*

The results therefore compare the fine-grid reference, the uncorrected coarse solver, and RECAST. In RECAST, SHRED-delta is embedded within the coarse-solver loop, and SHRED-SR reconstructs the corresponding fine-grid state according to the workflow in Figure 2.

### 3.1. Comparison with $P^2C^2$Net

Before presenting the results across the full PDE set, we compare RECAST with $P^2C^2$Net, a recent PDE-preserved coarse-correction network for spatiotemporal prediction [24]. $P^2C^2$Net is selected because it improves coarse-grid prediction while retaining an explicit PDE-based update structure and has been reported to outperform several neural and physics-informed baselines [24]. The comparison assesses whether RECAST achieves competitive or improved online prediction accuracy relative to a contemporary coarse-correction architecture.

For this comparison, $P^2C^2$Net is adapted to the same one-dimensional advection–diffusion and KdV–Burgers settings and is trained using the same coarse-grid data splits as RECAST. The implementation choices follow the original $P^2C^2$Net design as closely as possible after adaptation to the present one-dimensional scalar PDE setting. $P^2C^2$Net does not apply a learned correction to a separately executed coarse-grid solver. Instead, it recursively advances the coarse-grid state through an internal PDE-preserved update. In our one-dimensional adaptation, the correction at each step is performed by a Fourier neural operator that maps the current coarse-grid state to a corrected representation using 12 feature channels, 10 retained Fourier modes per spectral layer, and four spectral layers. The spatial derivatives required by each governing coarse-grid PDE are evaluated using learnable constrained five-point finite-difference filters, preserving the PDE-structured character of the original $P^2C^2$Net. In the advection–diffusion case, these filters represent the derivative terms associated with the advective and diffusive operators, whereas in the KdV–Burgers case, they include the nonlinear advective, viscous, and dispersive derivative terms. The resulting PDE-structured right-hand side is advanced with a fourth-order Runge-Kutta update, consistent with the time-stepping structure of the original $P^2C^2$Net.

$P^2C^2$Net is trained as a recursive coarse-grid sequence predictor. Following its own strategy, the model is trained in the coarse space as an autoregressive rollout predictor. The fine-grid reference trajectories are first block-averaged to the coarse grid, and the model learns to predict a sequence of future coarse-grid states from an initial coarse state using mean-squared error over the predicted rollout. Importantly, $P^2C^2$Net does not perform super-resolution: its output has the same spatial resolution as its coarse-grid input. Therefore, during evaluation, the $P^2C^2$Net rollout is mapped to the fine grid by nearest-neighbor upsampling for direct comparison with the fine-grid reference. In addition to the raw upsampled $P^2C^2$Net trajectory, we also report a diagnostic variant, in which the $P^2C^2$Net coarse rollout is passed through the trained offline SHRED-SR. This does not alter the $P^2C^2$Net dynamics and is not used during its rollout; it is included only to distinguish errors due to the coarse trajectory itself from errors due to the coarse-to-fine reconstruction step.

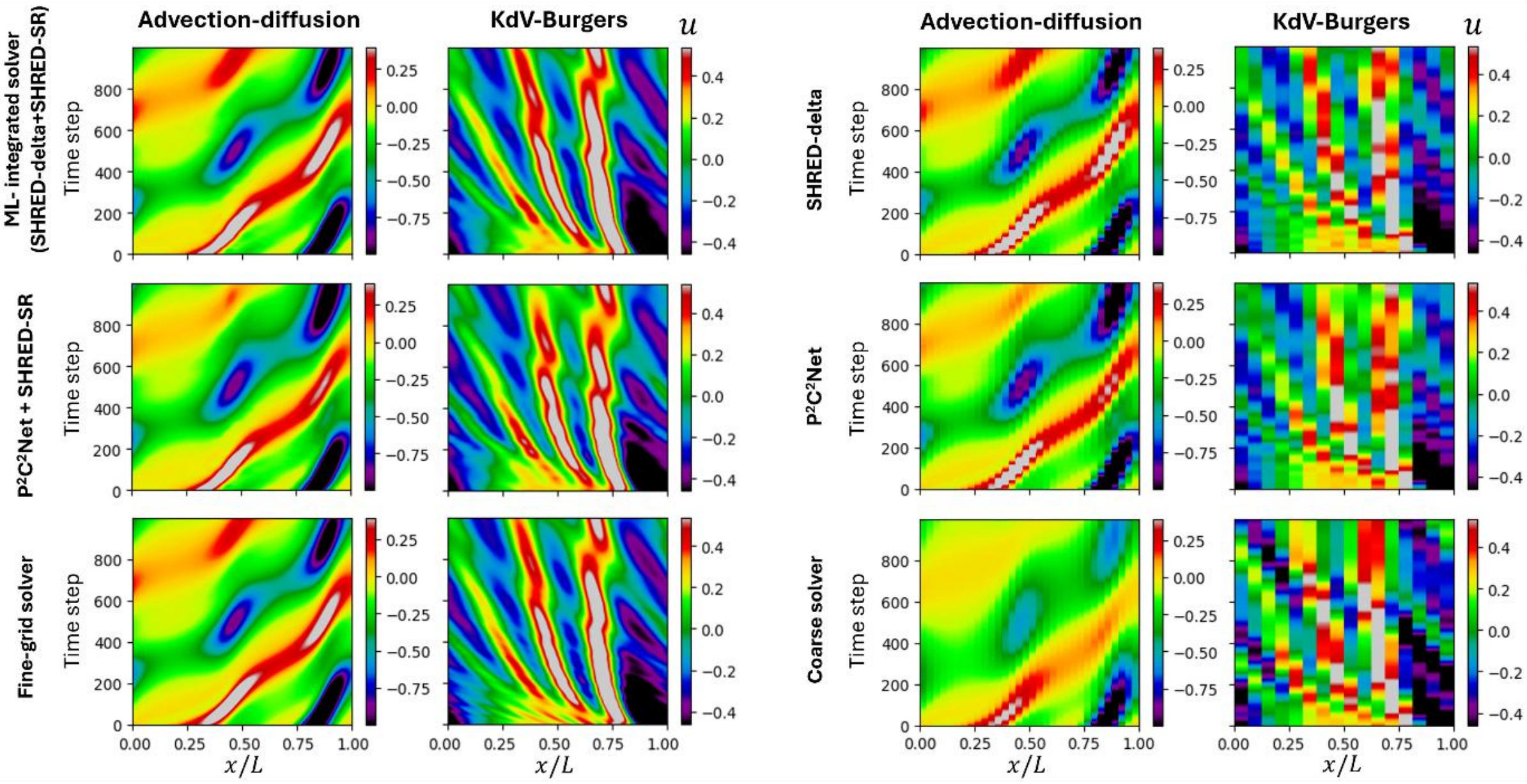

*Figure 3: Comparison of spatiotemporal solutions for advection-diffusion and KdV-Burgers test trajectories. The left group compares fine-grid fields obtained from RECAST, $P^2C^2$Net followed by offline SHRED-SR, and the fine-grid reference solver. The right group compares the corresponding coarse-grid trajectories from SHRED-delta, $P^2C^2$Net, and the pure coarse solver.*

The comparison is performed under intentionally demanding coarse-grid conditions. In particular, the coarse grid is chosen to be very under-resolved, especially for the KdV–Burgers equation, in order to test both approaches near their limiting regimes while maximizing the potential computational saving of the solver. This setting assesses each model's ability to maintain solver stability and a dynamically accurate trajectory over a long recursive rollout.

The sample spatiotemporal predictions (Figure 3), spatial profiles (Figure 4), and error curves (Figure 5) show that RECAST remains closer to the fine-grid reference than P²C²Net for both advection–diffusion and KdV–Burgers. P²C²Net improves upon the uncorrected coarse solver, but its trajectory accumulates phase, amplitude, and structural errors more rapidly than RECAST. Applying SHRED-SR to the P²C²Net rollout improves the spatial appearance of the reconstructed field but does not alter the underlying P²C²Net trajectory error.

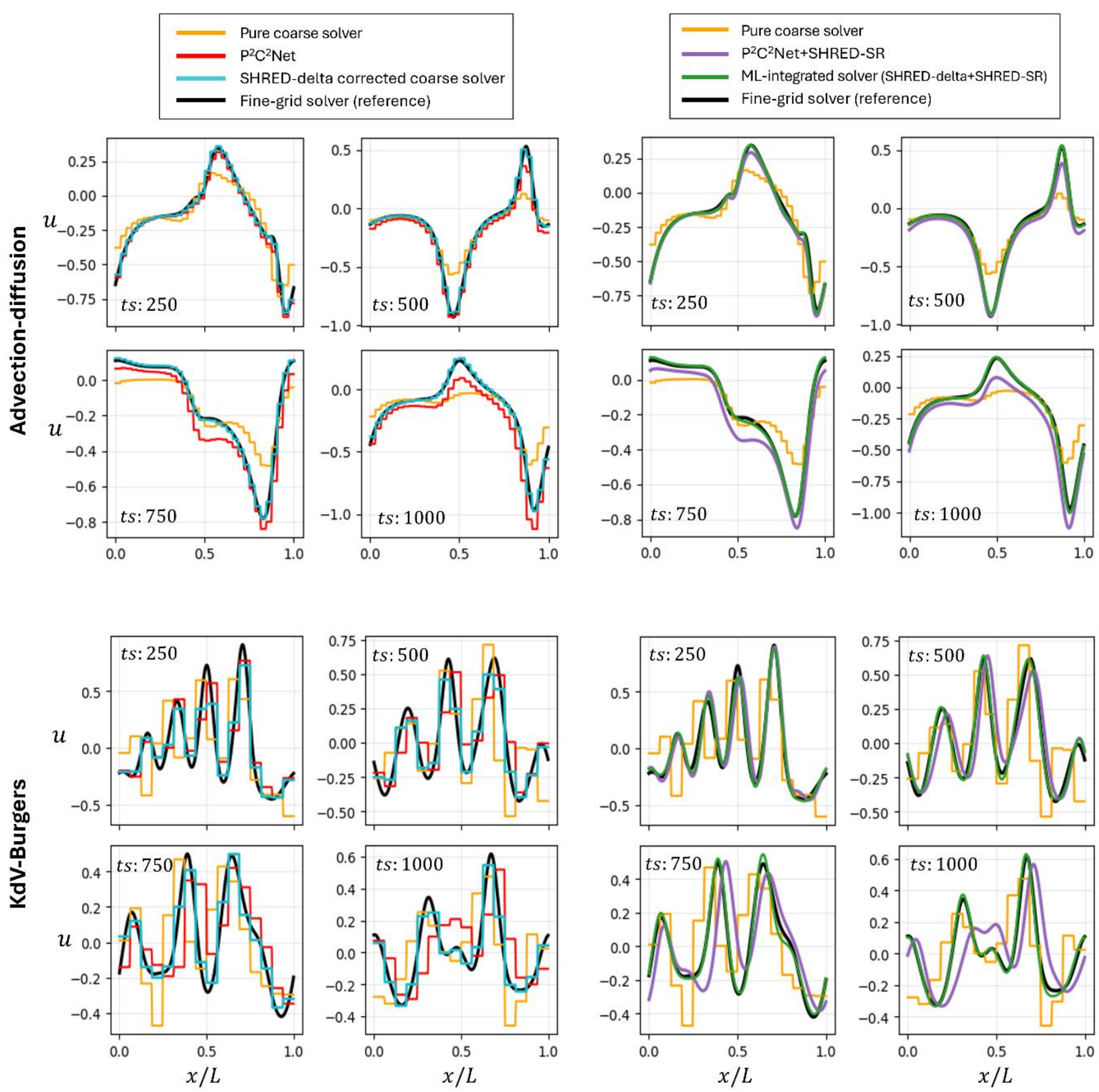

*Figure 4: Comparison of predicted spatial profiles at selected time steps for advection-diffusion and KdV-Burgers. Profiles are shown for the pure coarse solver, P²C²Net, SHRED-delta corrected coarse solver, P²C²Net followed by offline SHRED-SR, RECAST, and the fine-grid reference solver.*

The error curves in Figure 5 present a quantitative view of the observations in Figure 3 and Figure 4. For advection–diffusion, the RECAST and P²C²Net errors are comparatively close during the early rollout, after which the P²C²Net error grows steadily. For KdV–Burgers, the RECAST error is already lower during the early rollout, and the separation increases over time. For advection–diffusion, the time-averaged mean relative error over the 1000-step rollout is $4.69 \times 10^{-1}$ for the pure coarse-grid solver, $2.09 \times 10^{-1}$ for P²C²Net, $1.54 \times 10^{-1}$ for P²C²Net+SHRED-SR, $1.35 \times 10^{-1}$ for the SHRED-delta corrected coarse solver, and $6.83 \times 10^{-2}$ for RECAST. Thus, RECAST reduces the error by about 67% relative to P²C²Net and by about 56% relative to P²C²Net+SHRED-SR.

For KdV–Burgers, the corresponding errors are $6.36 \times 10^{-1}$, $5.44 \times 10^{-1}$, $4.27 \times 10^{-1}$, $3.58 \times 10^{-1}$, and $1.59 \times 10^{-1}$, respectively. In this more difficult case, RECAST reduces the error by about 71% relative to P²C²Net and about 63% relative to P²C²Net+SHRED-SR.

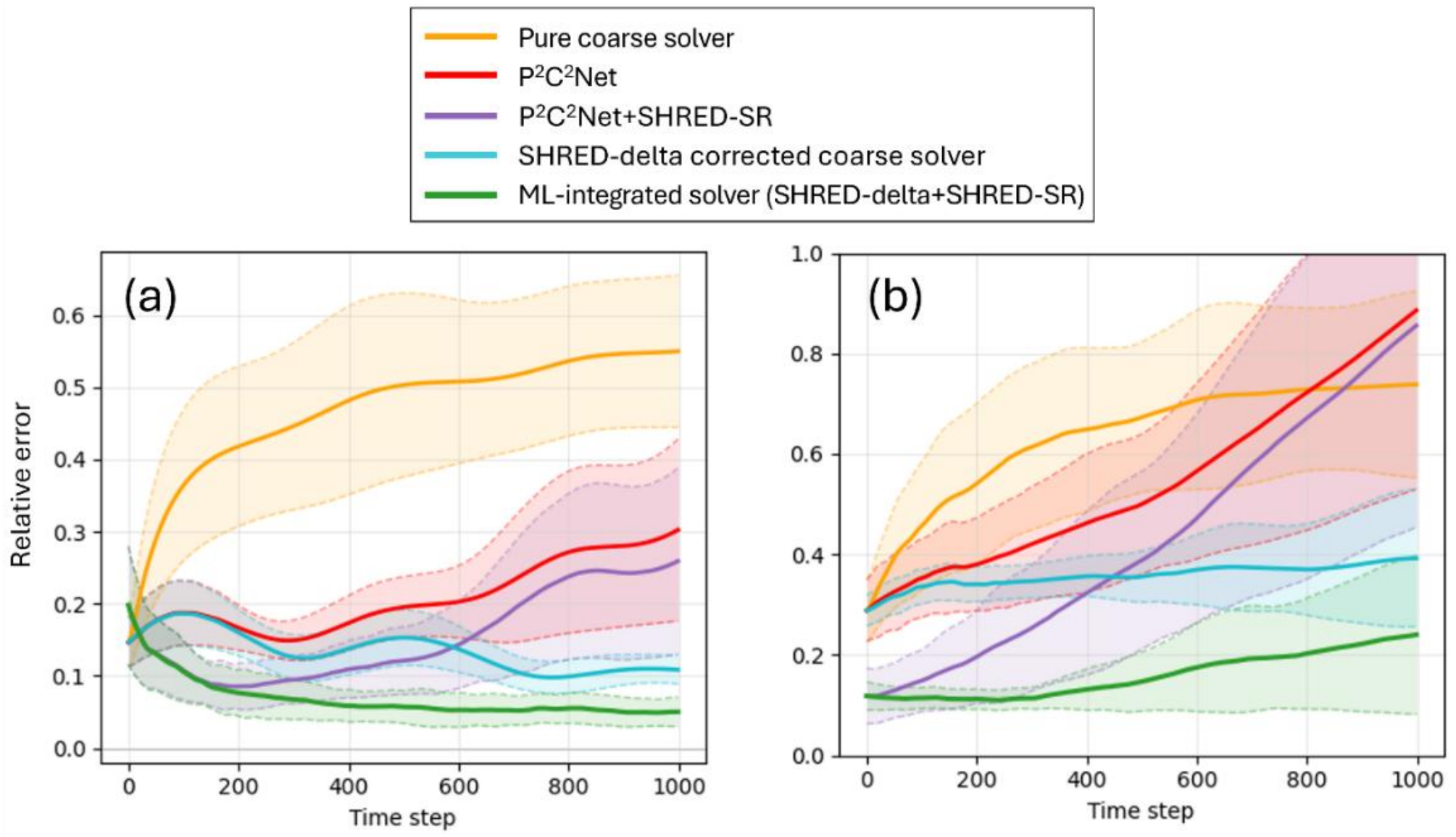

*Figure 5: Mean relative error evolution over time steps for (a) advection-diffusion, and (b) KdV-Burgers. Errors are shown for the pure coarse solver, $P^2C^2$Net, $P^2C^2$Net followed by offline SHRED-SR, the SHRED-delta corrected coarse solver, and the full ML-integrated solver (RECAST). Mean values are calculated over all test trajectories, and the shaded bands indicate $\pm 1$ standard deviation across the test trajectories.*

In Appendix A, the comparison is extended to a 5000-step rollout to examine longer-horizon error accumulation. In the extended rollout, the $P^2C^2$Net prediction degrades substantially for KdV–Burgers and becomes unstable for advection–diffusion, whereas RECAST continues to track the fine-grid reference more closely. Also, in both PDE cases, the $P^2C^2$Net error reaches and then exceeds the pure coarse-grid-solver error around, or shortly after time step 1000, which is approximately the trajectory length used during training, although the test rollouts start from unseen initial conditions. Beyond outperforming $P^2C^2$Net in these cases, RECAST has an additional advantage in this highly coarse setting because the combination of online trajectory correction by SHRED-delta and subsequent super-resolution by SHRED-SR addresses both evolution error and representation error over long prediction horizons. This is whereas $P^2C^2$Net does improve the coarse trajectory dynamics but does not natively reconstruct the corresponding fine-grid state.

**3.2. Closed-loop performance across the PDE test cases**

The spatiotemporal results in Figure 6 to Figure 9 demonstrate that applying the learned correction at each time step substantially reduces accumulated evolution error, allowing both the dynamics and reconstructed spatial details produced by RECAST to remain close to the reference over long rollouts. Figure 10 further confirms this behavior through spatial profiles at representative time steps.

Across the six PDE systems, RECAST reduces the dominant case-dependent coarse-grid error mechanisms and preserves the physically relevant dynamical structures. For advection–diffusion, the correction mainly limits excessive smoothing and diffusive spreading, preserving the width and amplitude of the transported concentration bands. For KdV–Burgers, it restores the smooth nonlinear wave profile, recovering peak and trough amplitudes together with phase-aligned oscillations. For the dielectric wave problem, RECAST preserves coherent electric-field wave fronts through the dielectric region, suppressing the rapid damping and numerical dispersion observed in the pure coarse-grid solver.

In the FitzHugh–Nagumo system, the correction preserves traveling excitation-recovery structures by preventing the suppression and displacement of activation pulses caused by the coarse-grid solver, hence maintaining the timing and spatial extent of the excited and refractory regions. For the Schrödinger equation, RECAST corrects phase accumulation and interference, preserving the probability-density pattern and phase structure, both of which are strongly distorted by the uncorrected coarse-grid solver. For the shallow-water system, the correction recovers propagating wave trains, reflection patterns, and oscillation phase over topography, whereas the pure coarse-grid rollout severely distorts the phase and collapses much of the height-field evolution into nearly stationary bands.

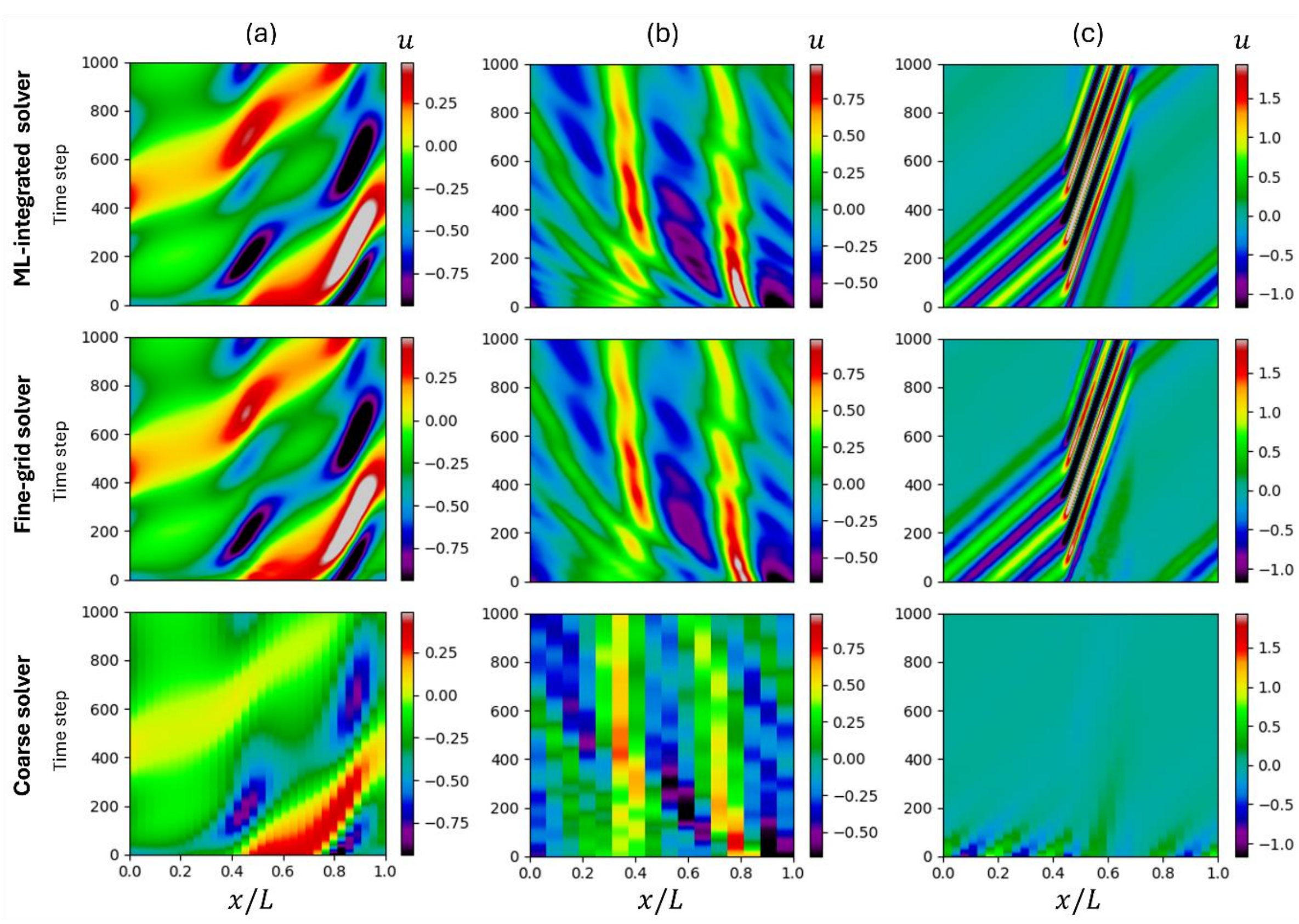

*Figure 6: Comparison of spatiotemporal solutions for test trajectories obtained from RECAST, fine-grid reference solver, and pure coarse-grid solver for the PDE cases: (a) advection-diffusion, (b) KdV-Burgers, and (c) electric-field propagation in an inhomogeneous dielectric.*

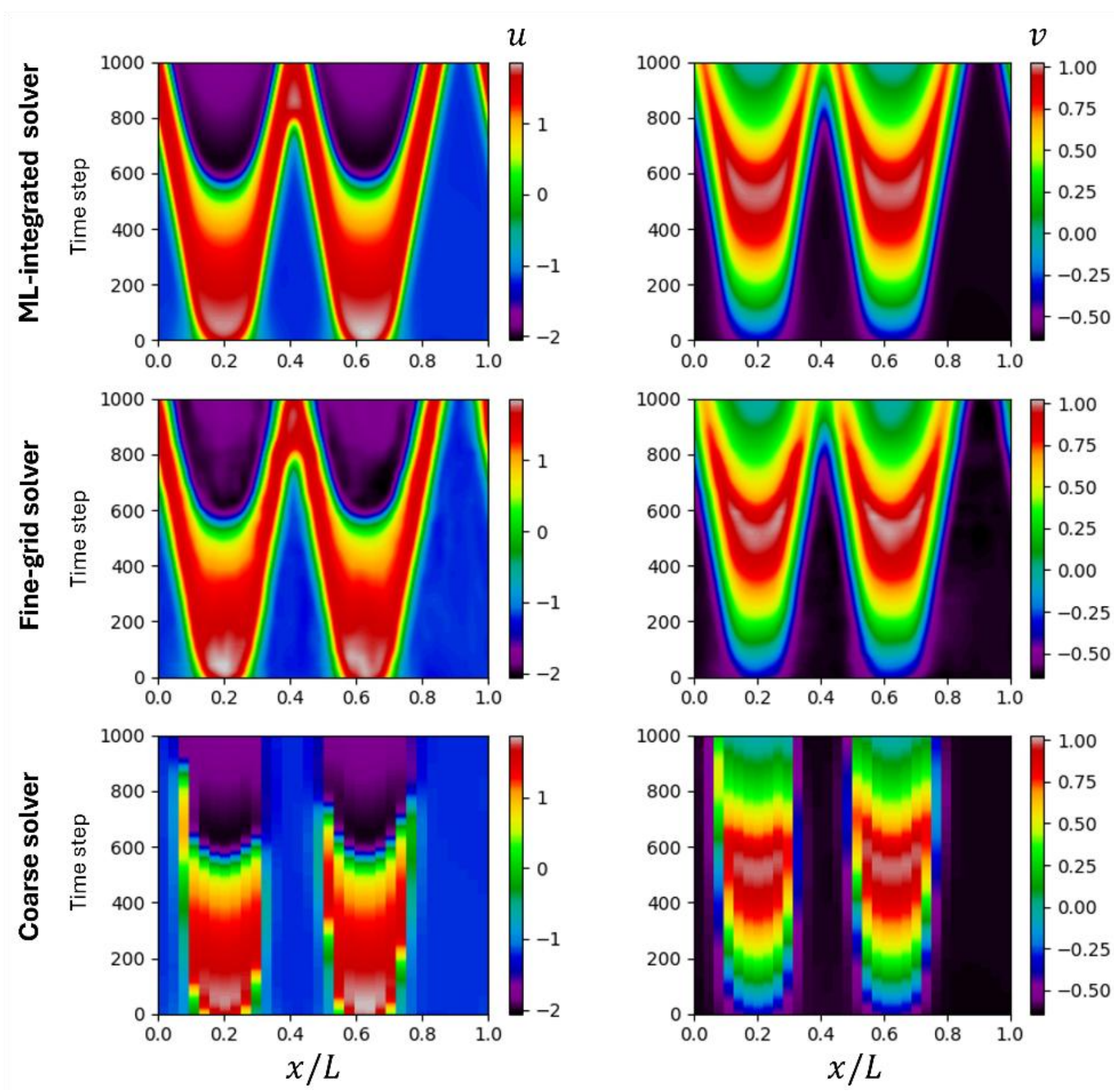

*Figure 7: Comparison of spatiotemporal solutions for a test trajectory obtained from the ML-integrated solver (RECAST), fine-grid reference solver, and pure coarse-grid solver for the FitzHugh-Nagumo equation.*

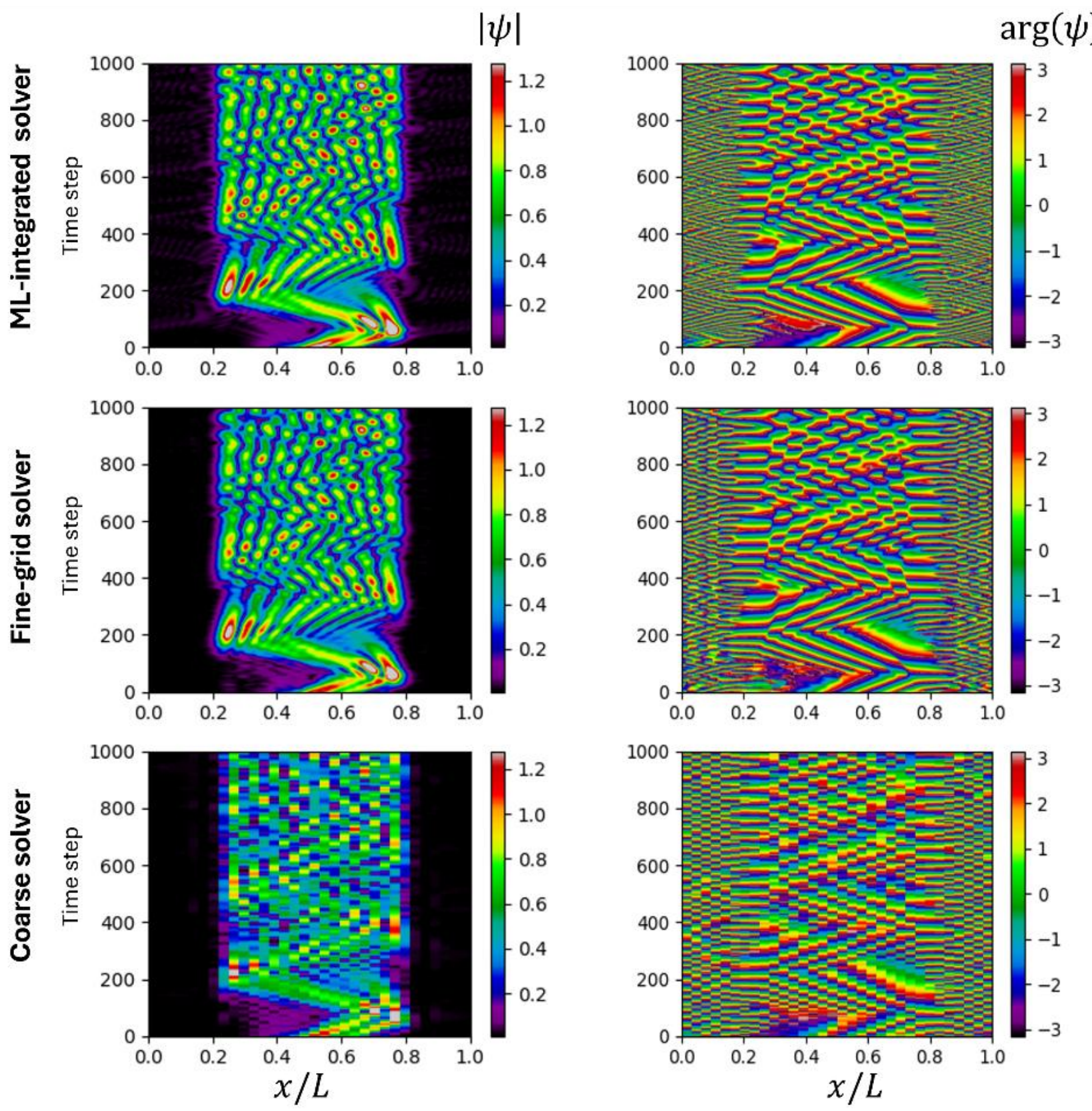

*Figure 8: Comparison of spatiotemporal solutions for a test trajectory obtained from RECAST, fine-grid reference solver, and pure coarse solver for the Schrödinger equation.*

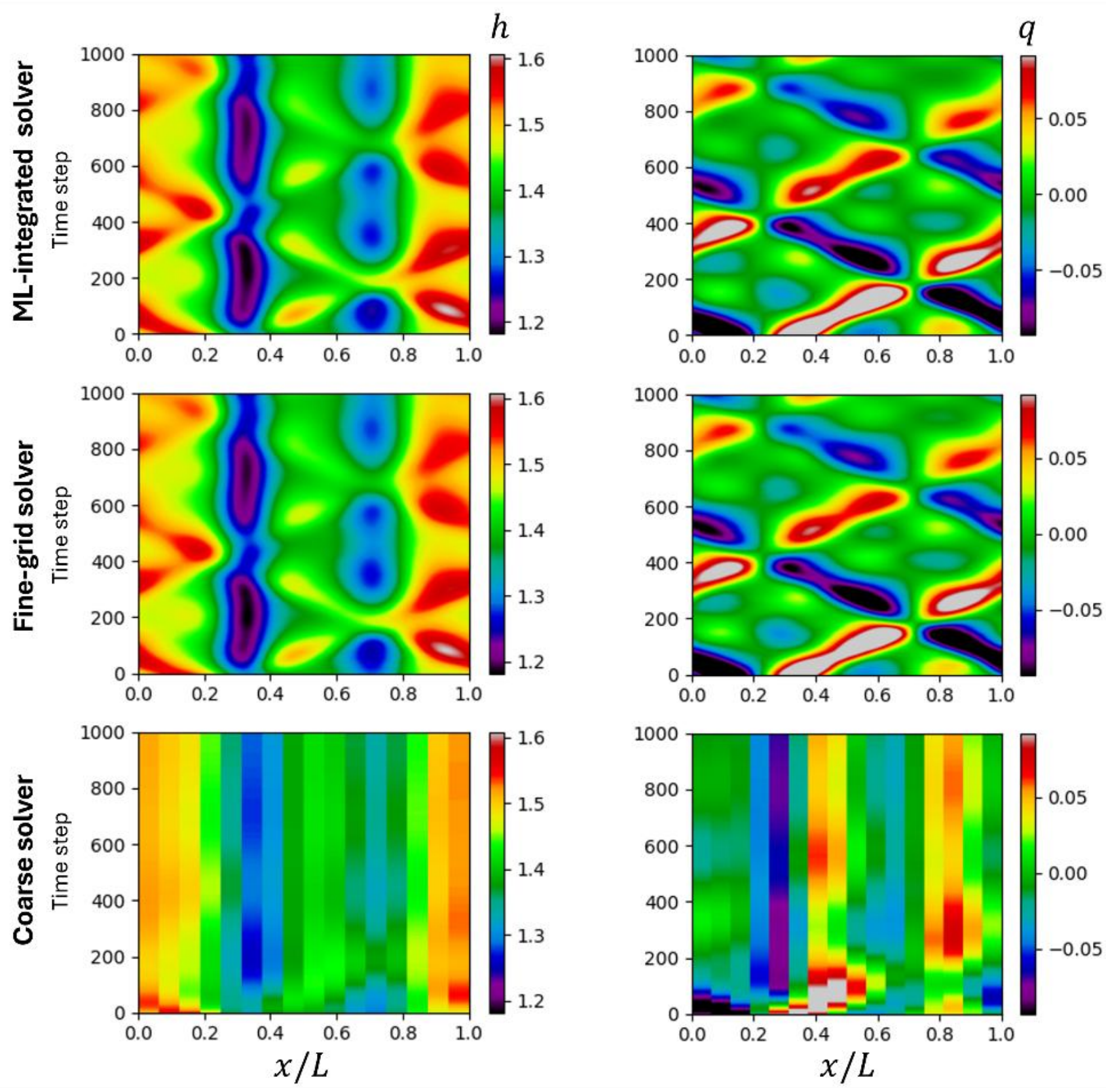

*Figure 9: Comparison of spatiotemporal solutions for a test trajectory obtained from RECAST, fine-grid reference solver, and pure coarse-grid solver for the shallow-water equation.*

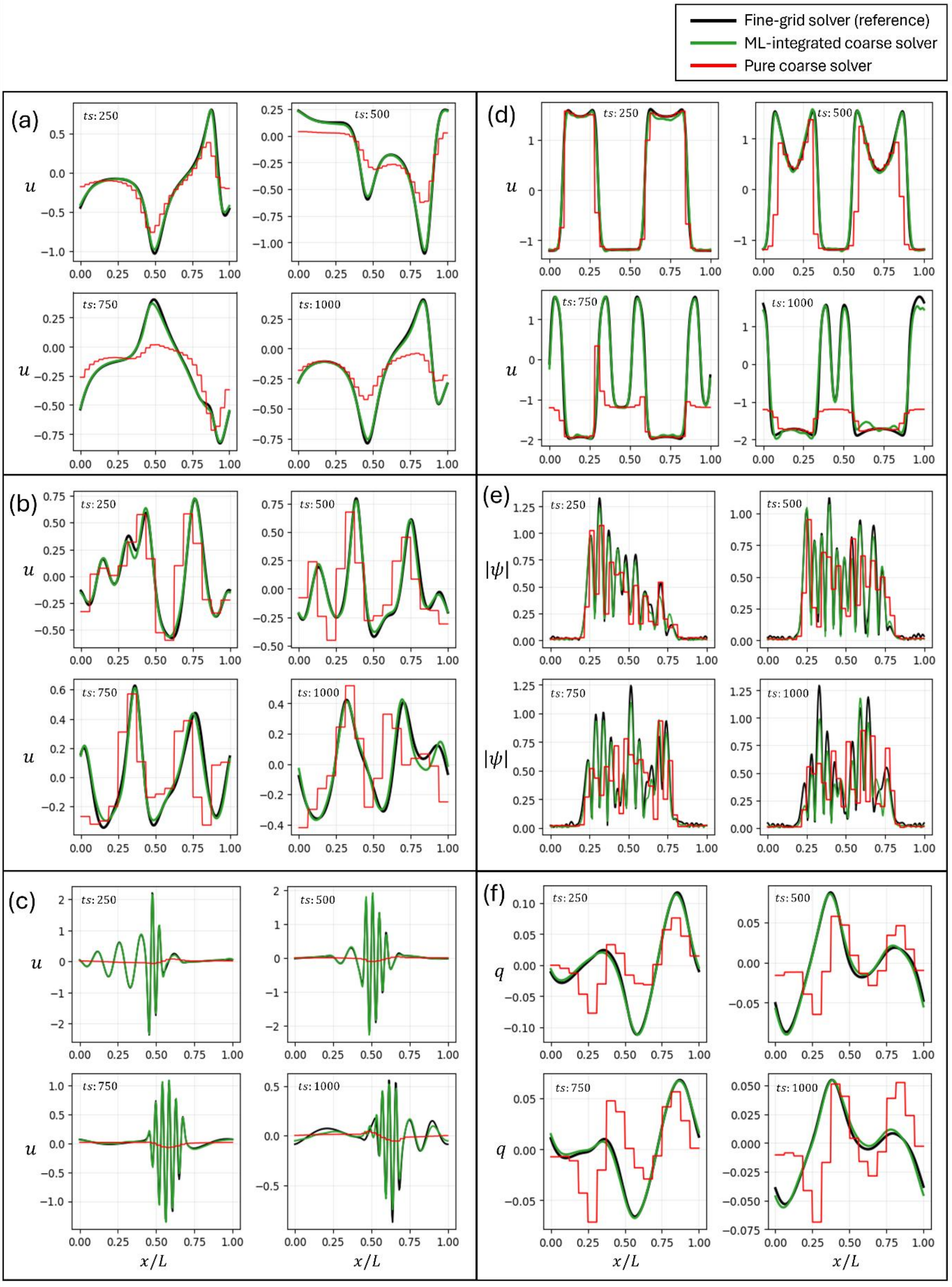

*Figure 10: Comparison of spatial profiles at various time steps from the test trajectories, obtained from RECAST, fine-grid reference solver, and pure coarse solver for the PDE cases: (a) advection-diffusion, (b) KdV-Burgers, and (c) electric-field propagation in an inhomogeneous dielectric, (d) FitzHugh-Nagumo, (e) Schrödinger, (f) shallow-water equation.*

Overall, these results indicate that RECAST limits long-time trajectory drift. This is the role of SHRED-delta, which acts inside the time-stepping loop by correcting each provisional coarse-grid update before it is used as the input to the next solver step. As a result, SHRED-delta serves as a learned closure-resembling correction between resolved coarse variables and unresolved fine-grid effects. Its strongest use cases are PDE regimes in which sub-grid scales influence the larger-scale dynamics.

The recovery of fine-grid spatial detail is then provided by SHRED-SR, which maps from the dynamically corrected coarse trajectory to the corresponding high-resolution state. The results show that, across the test cases, the reconstructed fields retain fine-scale features that are largely absent from the coarse representation, including sharp gradients, localized extrema, coherent wave fronts, and small-scale oscillatory structures. The agreement between RECAST prediction and the fine-grid reference demonstrates the combined capability of SHRED-delta for trajectory stabilization and SHRED-SR for fine-resolution spatial reconstruction, underlining their complementary roles within the framework.

Additional solution examples for held-out advection–diffusion and KdV–Burgers test trajectories with different unseen initial-condition realizations are provided in Appendix B, which is meant to further demonstrate robustness across different initial conditions.

The error plots in Figure 11 and Figure 12 quantify the closed-loop rollout comparison shown visually across the tested PDEs. For test trajectories $j$, the instantaneous relative error at time step $n$ is defined as

$$\varepsilon_j(n) = \frac{\left\| u_j^n - \hat{u}_j^n \right\|_2}{\left\| u_j^n \right\|_2}, \qquad j = 1, 2, \ldots, N_{test}, \tag{Eq. 30}$$

where $u_j^n$ is the fine-grid reference solution, and $\hat{u}_j^n$ denotes the prediction from either the ML-integrated solver (RECAST), the pure coarse-grid solver, or the offline SHRED-SR reconstruction of the uncorrected coarse trajectory. The mean error plotted over time (Figure 11) is

$$\bar{\varepsilon}(n) = \frac{1}{N_{\text{test}}} \sum_{j=1}^{N_{\text{test}}} \varepsilon_j\,(n), \tag{Eq. 31}$$

with the shaded bands representing $\pm 1$ standard deviation across test trajectories. Figure 12 plots the time-averaged error for each trajectory over $N_t$ rollout time steps

$$\langle \varepsilon_j \rangle_t = \frac{1}{N_t} \sum_{n=1}^{N_t} \varepsilon_j\,(n). \tag{Eq. 32}$$

Figure 11 therefore shows the accumulated error of the pure coarse solver, and the improvement from ML correction and super-resolution. In addition, the plots show applying SHRED-SR only as an offline reconstruction of this uncorrected coarse trajectory gives errors of comparable magnitude as in pure coarse solver in most cases. This is expected as the super-resolution alone cannot recover an accurate fine-grid solution once the underlying coarse dynamics have already drifted. In contrast, the ML-integrated solver maintains consistently lower errors because SHRED-delta corrects the coarse state inside the time-stepping loop before SHRED-SR reconstructs the fine-grid field.

The time-averaged error plot in Figure 12 shows the mean accumulated rollout error over the 1000-step test trajectories. Across the six PDE cases, the pure coarse-grid solver has approximately $2\times$ to $13\times$ the time-averaged relative error of RECAST over the 1000-step test trajectories, corresponding to approximately 50% to 92% lower error for RECAST, while RECAST evolves on grids with only 32 or 16 coarse-grid points rather than 256 fine-grid points, corresponding to $8\times$ and $16\times$ spatial coarsening. This is a demanding setting because the comparison is made after long recursive rollouts and the error is measured against much finer grid reference solution.

The error of RECAST reflects two coupled contributions: the ability of SHRED-delta to keep the coarse trajectory dynamically aligned with the fine-grid reference, and the ability of SHRED-SR to reconstruct the fine-grid states from the corrected coarse states. To isolate the reconstruction contribution, Figure 13 evaluates SHRED-SR offline using projected coarse-grid histories obtained by applying the block-averaging restriction of Section 2.1 to the fine-grid test trajectories. The error of the projected coarse-grid reference is compared with the SHRED-SR reconstruction error relative to the same fine-grid reference. Across the six PDE cases, SHRED-SR reduces the error of the projected coarse representation by factors of approximately $2\times$ to $9\times$. This confirms that SHRED-SR recovers fine-grid spatial information when supplied with dynamically consistent coarse histories.

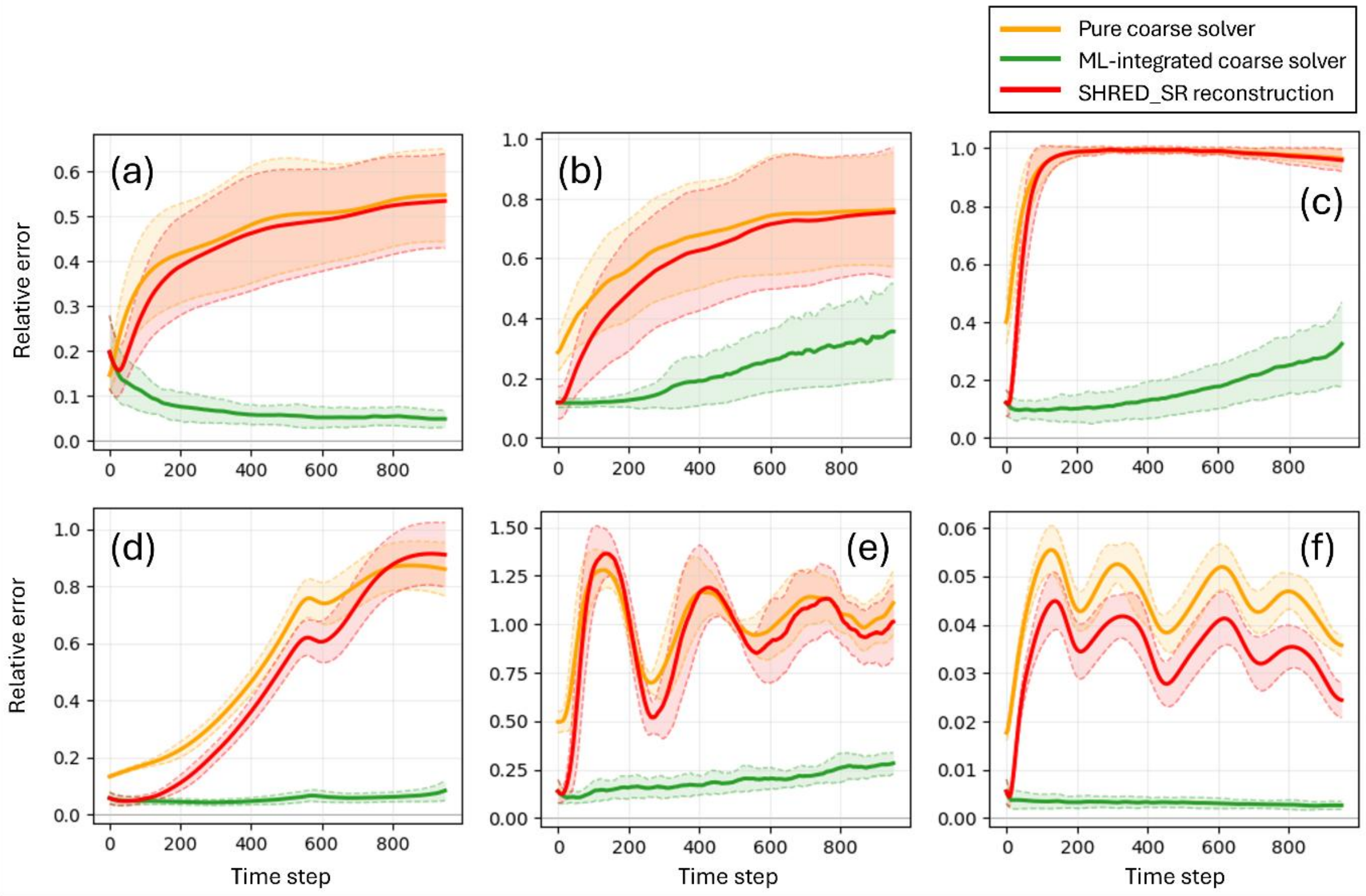

*Figure 11: Mean relative error (Eq. 31) evolution over time steps for the ML-integrated coarse solver (RECAST), pure coarse solver, and SHRED-SR fine-grid reconstruction of the uncorrected coarse-solver state (without SHRED-delta application), with respect to the fine-grid reference solver across the PDE cases: (a) advection-diffusion, (b) KdV-Burgers, and (c) Electric-field propagation in an inhomogeneous dielectric, (d) FitzHugh-Nagumo, (e) Schrödinger, (f) shallow-water equation. Mean values are calculated over all test trajectories, and shaded bands indicate $\pm 1$ standard deviation across the test trajectories.*

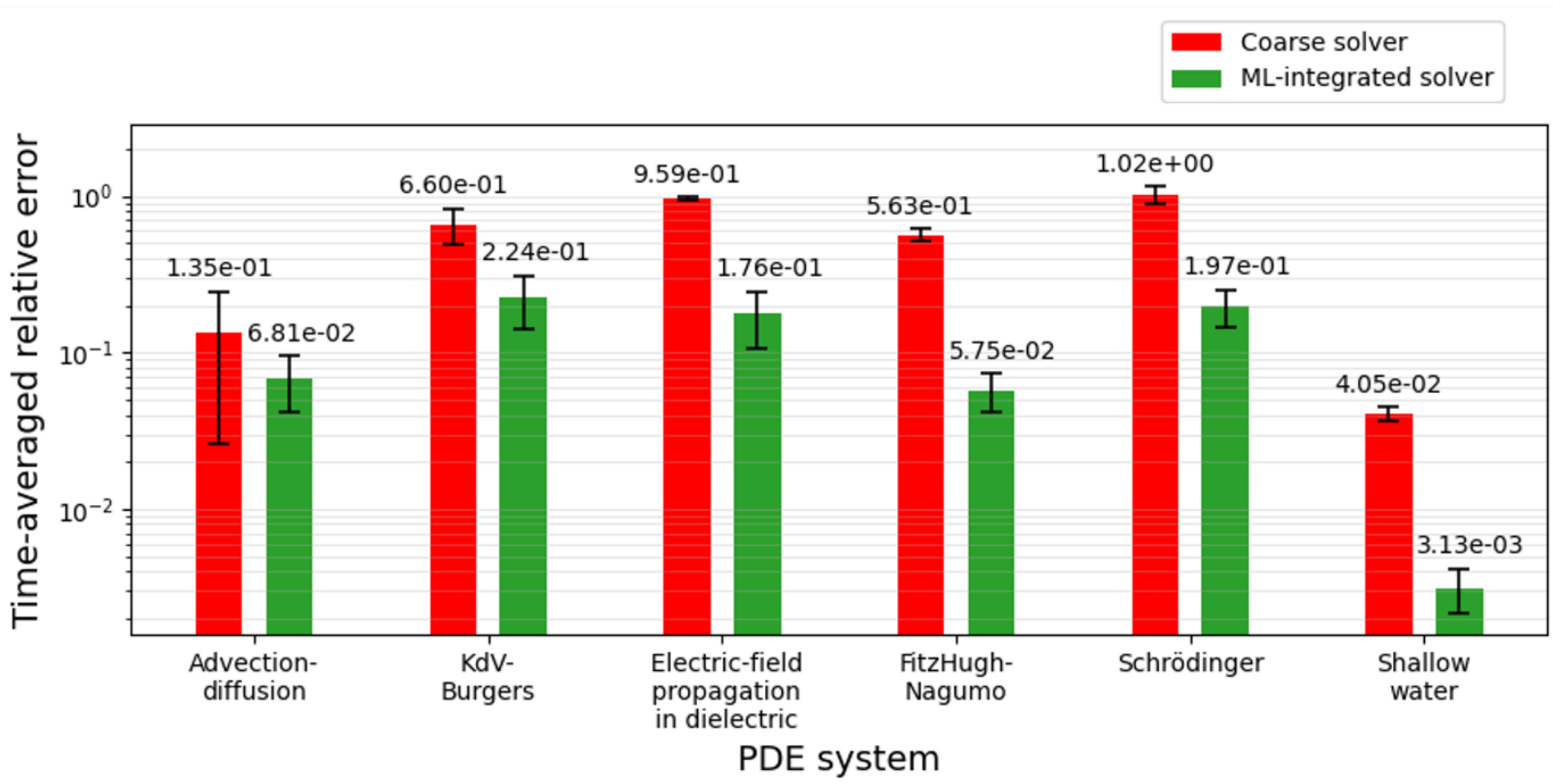

*Figure 12: Time-averaged mean relative error (Eq. 32) of RECAST and pure coarse-grid solver with respect to the fine-grid reference across different PDEs. Mean values are calculated over all test trajectories and error bars indicate $\pm 1$ standard deviation across test trajectories.*

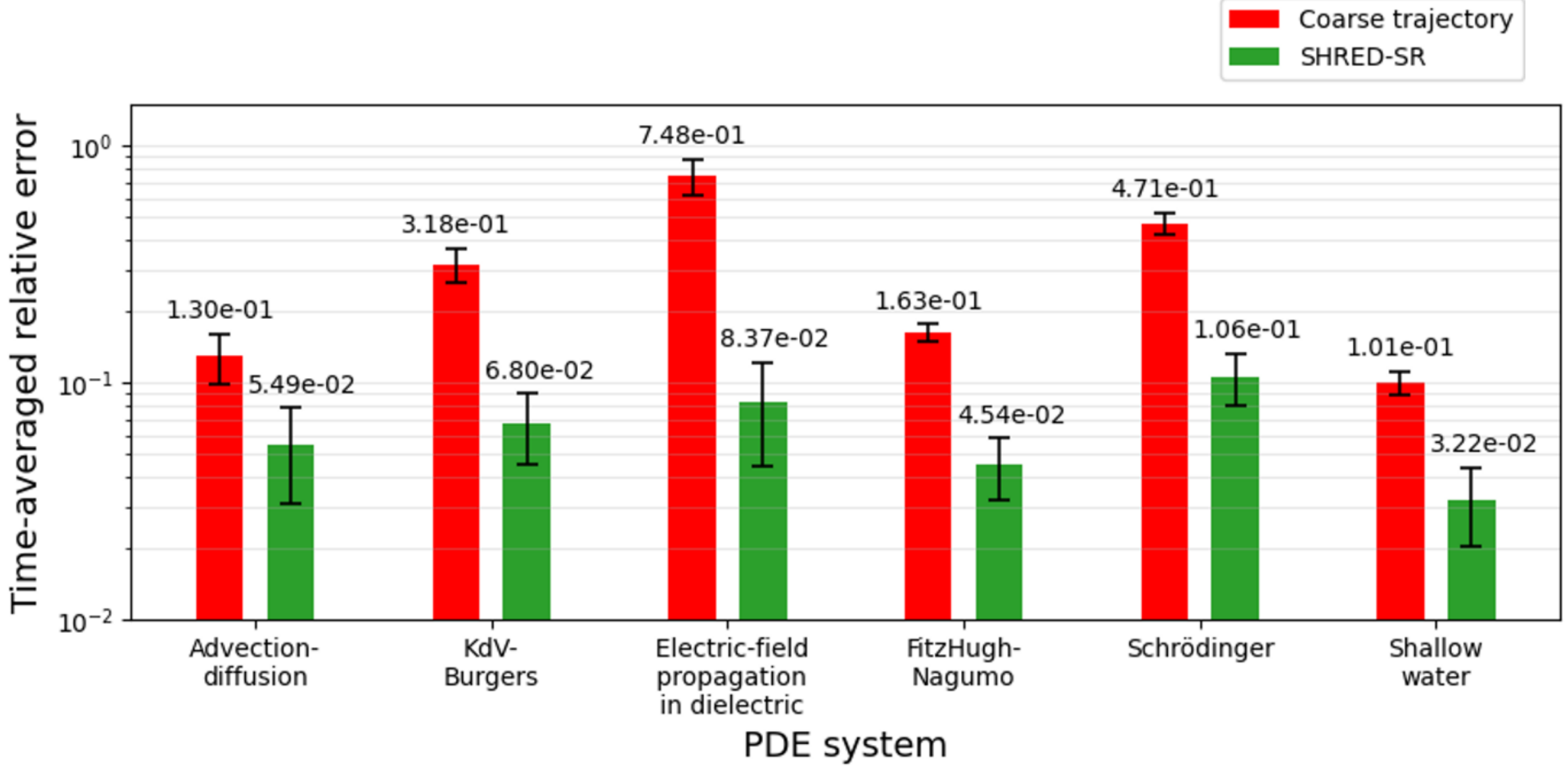

*Figure 13: Time-averaged mean relative error (Eq. 32) of the offline SHRED-SR reconstruction applied to the block-averaged coarse trajectory with respect to the fine-grid reference across different PDEs. Mean values are calculated over all test trajectories, and error bars indicate $\pm1$ standard deviation across test trajectories.*

### 3.3. Parametric generalization

To assess parameter-dependent generalization of RECAST, additional tests were performed in which the framework was trained over a prescribed parameter range and evaluated on held-out parameter realizations not used during training. These demonstrations are distinct from the previous initial-condition generalization results: here, the governing dynamics themselves vary through the PDE parameters. The tests are conducted for the advection–diffusion and KdV–Burgers equations. For advection–diffusion, $c_{\mathrm{adv}}$ is varied over $[0.3, 3.0]$. For the KdV–Burgers, the viscosity is varied over $\nu \in [0.01, 1.0]$.

The results in Figure 14 to Figure 16 show that RECAST solution remains aligned with the fine-grid reference across different parameter values in both PDEs. In the advection–diffusion case, increasing $c_{\mathrm{adv}}$ changes the advective transport rate and therefore the position, inclination, and spreading of the transported bands. The pure coarse-grid solver shows parameter-dependent displacement and smoothing, whereas RECAST preserves the correct advective speed and diffusion-controlled width across the tested values. This indicates that the learned correction is not tied to a single transport speed but adapts to how the coarse-grid error changes with the advection regime.

For KdV–Burgers, varying $\nu$ changes the balance between nonlinear steepening, dispersive oscillation, and viscous smoothing. The fidelity of the uncorrected coarse solver varies visibly across $\nu$. RECAST, however, preserves sharper oscillatory profiles at lower viscosity and recovers smoother, more dissipative evolution at higher viscosity, consistent with the corresponding fine-grid solutions.

Figure 17 shows that, in both cases, the uncorrected coarse-grid solver accumulates error rapidly and exhibits substantial variation across parameter regimes. RECAST, however, maintains low error throughout the rollout with much smaller variation.

The parameter-binned error plots in Figure 18 additionally quantify the error improvement across parameter ranges. Test trajectories are grouped into parameter bins, and the time-averaged relative error is averaged within each bin. For advection–diffusion, RECAST reduces error by approximately $4.5\times$ to $34\times$, corresponding to about 78% to 97% lower error across the tested $c_{\mathrm{adv}}$ bins. For KdV–Burgers, the error reduction by RECAST is approximately $9\times$ to $30\times$, corresponding to about 89% to 97% lower error across the tested viscosity bins.

These results suggest that the learned correction acts as a parameter-dependent closure that adapts to changes in the influence of unresolved scales on the coarse trajectory.

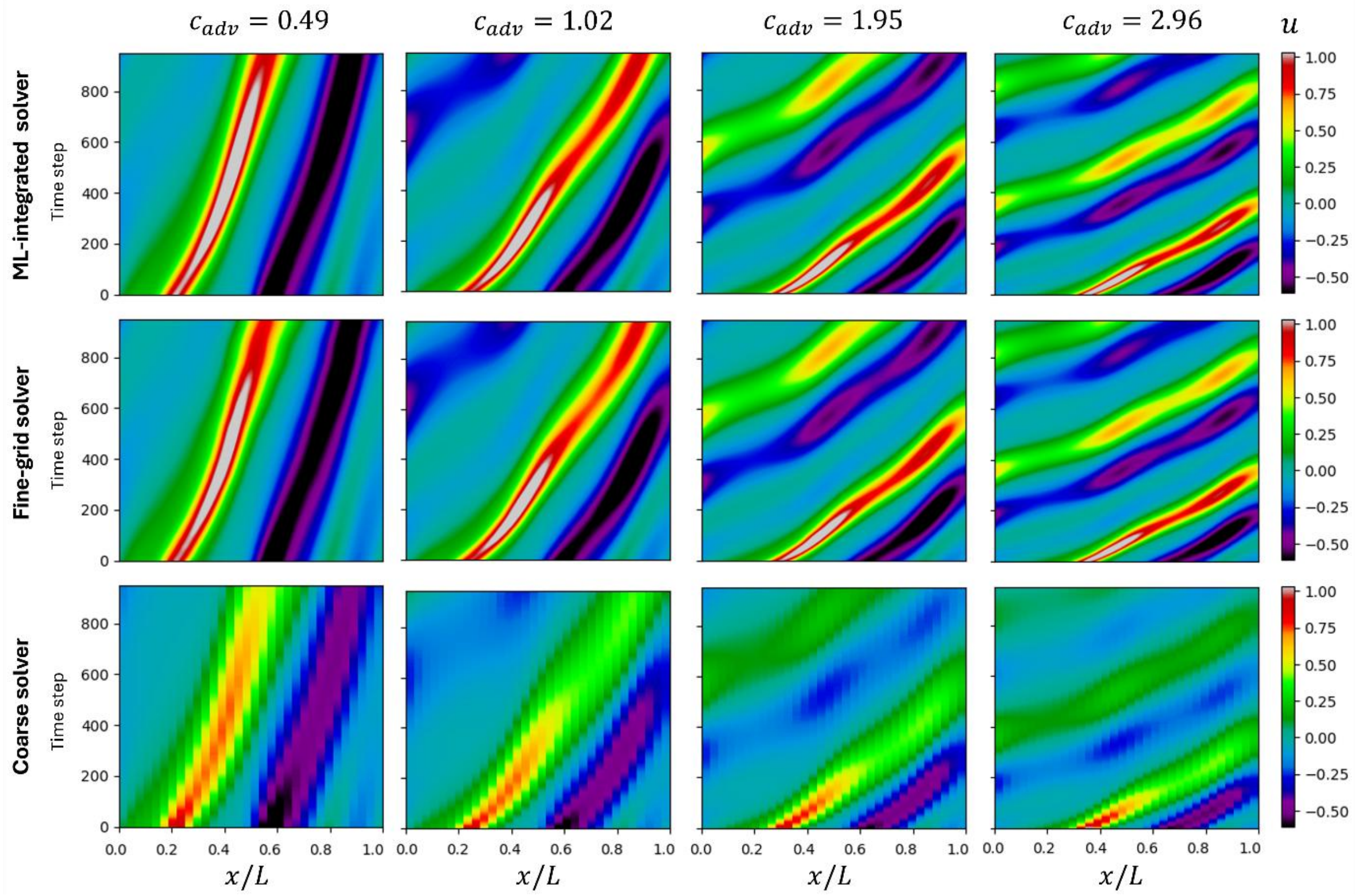

*Figure 14: Comparison of spatiotemporal solutions for different test parameter realizations ($c_{adv}$) of the advection-diffusion equation, obtained using RECAST, fine-grid reference solver, and pure coarse-grid solver.*

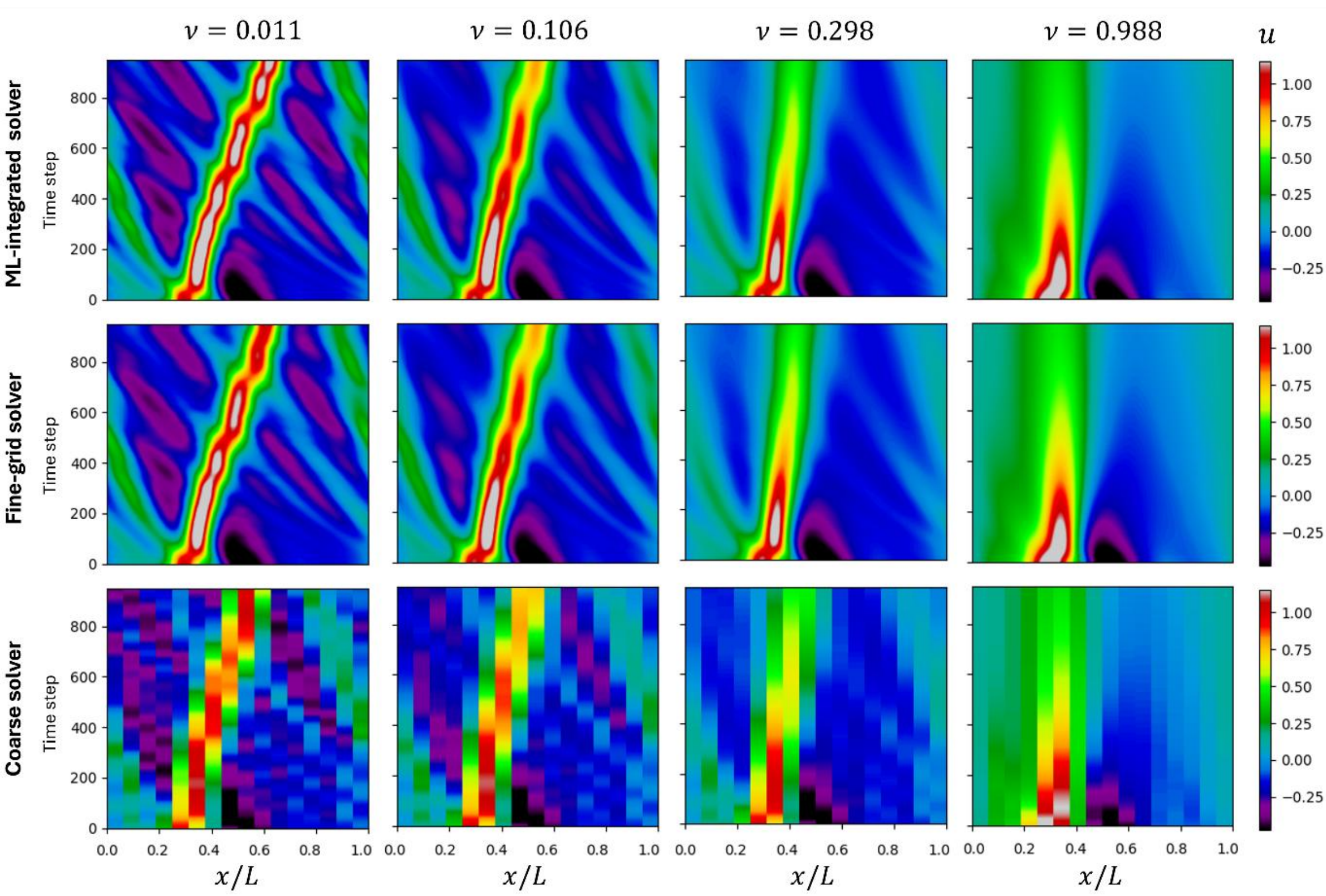

*Figure 15: Comparison of spatiotemporal solutions for different test parameter realizations ($\nu$) of the KdV-Burgers equation, obtained using RECAST, fine-grid reference solver, and pure coarse-grid solver.*

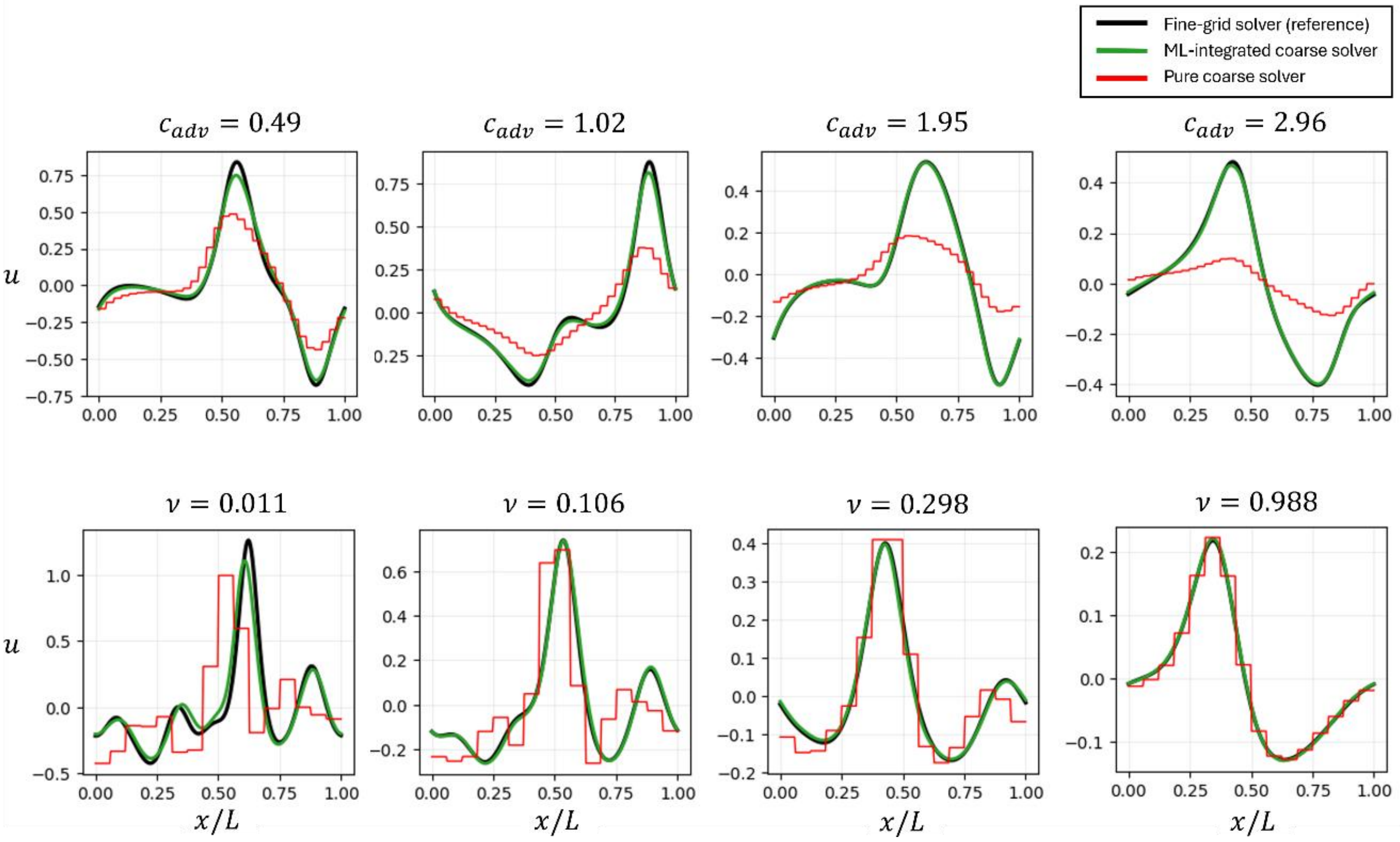

*Figure 16: Comparison of spatial profiles at time step 1000, obtained from RECAST, fine-grid reference solver, and pure coarse-grid solver for different test parameter realizations of advection-diffusion (upper row) and KdV-Burgers (lower row).*

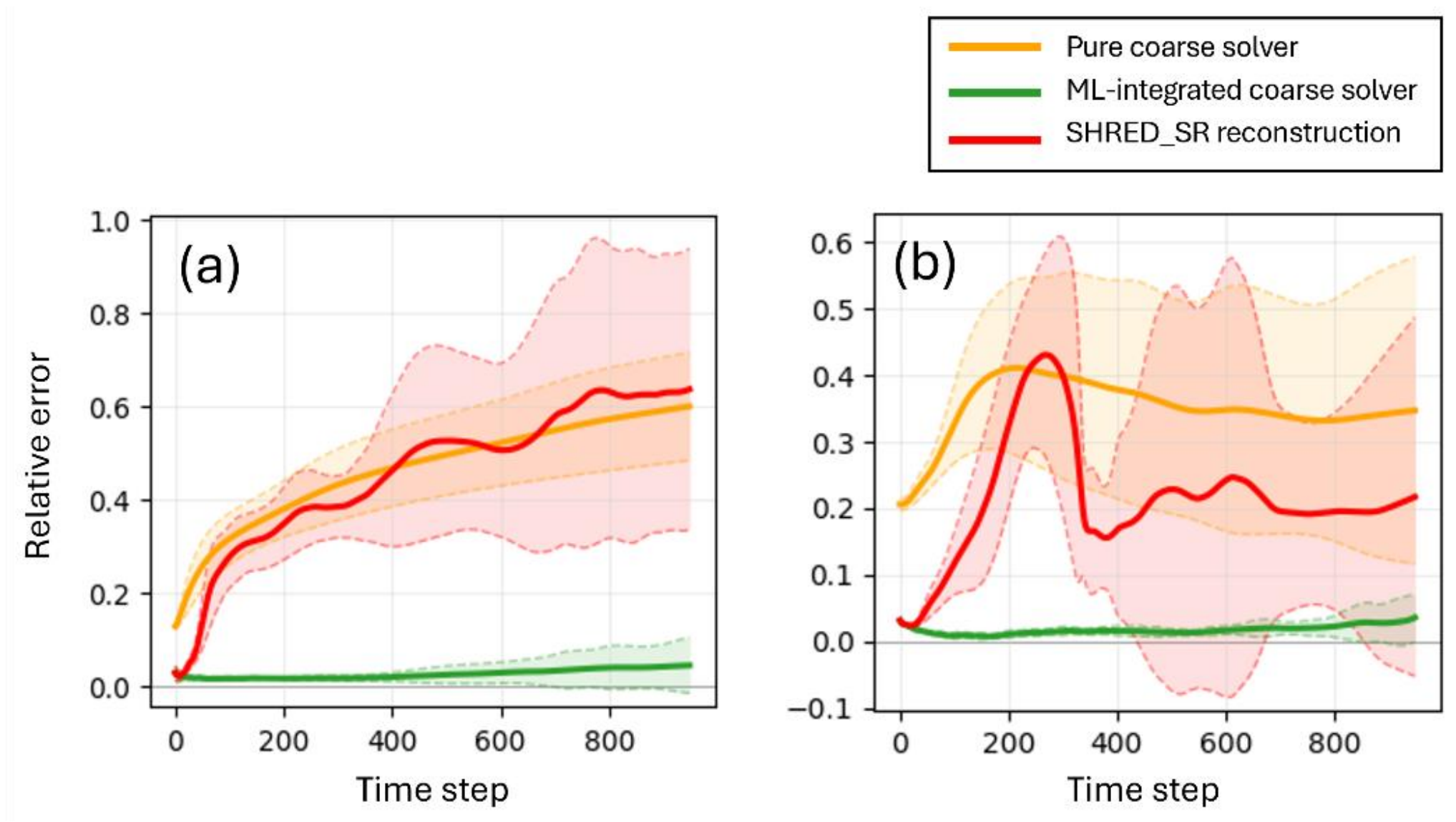

*Figure 17: Mean relative error (Eq. 31) evolution over time steps for RECAST, pure coarse-grid solver, and SHRED-SR fine-grid reconstruction of the uncorrected coarse-solver state (without SHRED-delta application), with respect to the fine-grid reference solver for (a) advection-diffusion and (b) KdV-Burgers. Mean values are calculated over all test trajectories and shaded bands indicate $\pm 1$ standard deviation across test trajectories.*

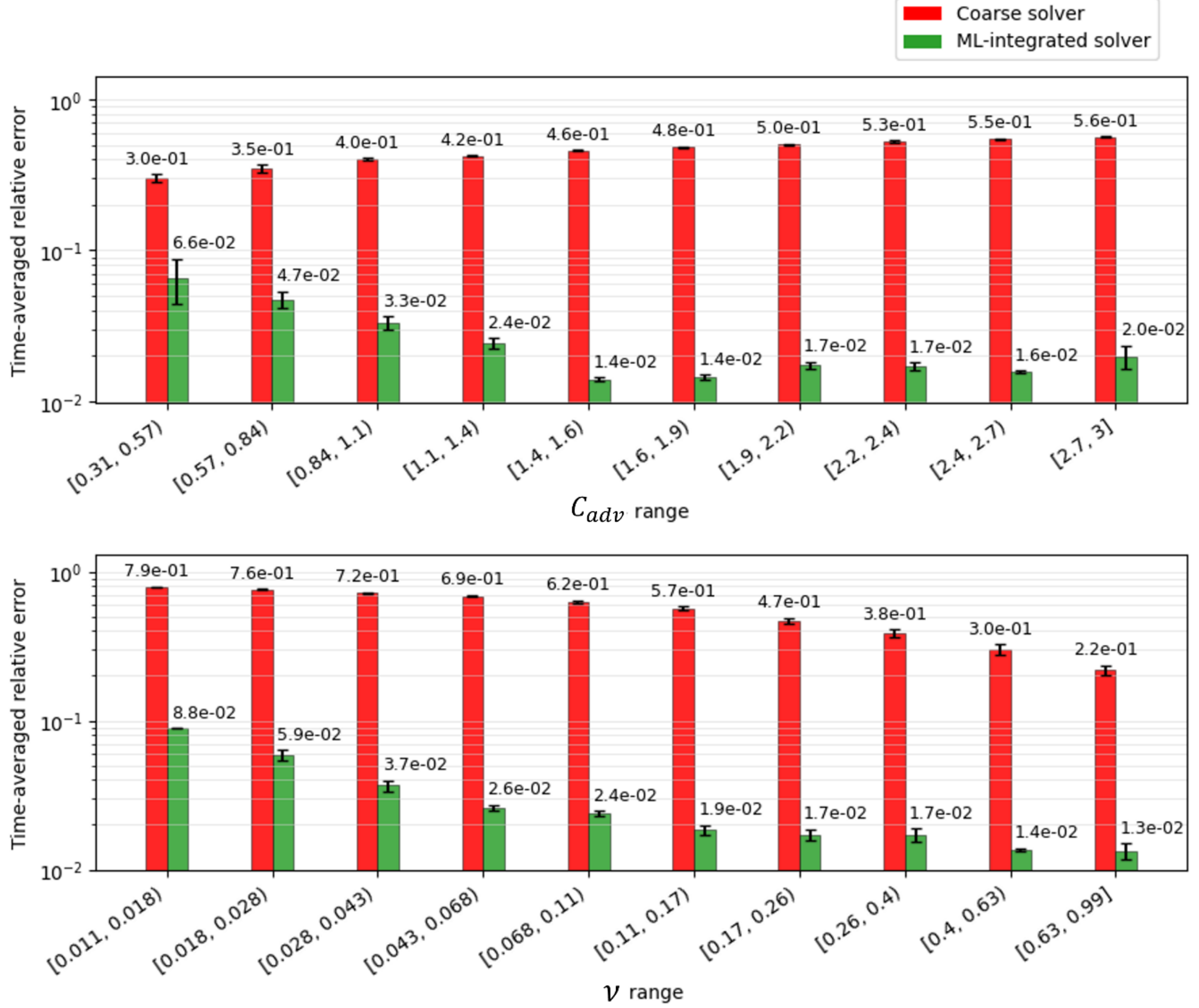

*Figure 18: Variation of time-averaged relative error (Eq. 32) of RECAST and pure coarse-grid solver across parameter ranges for $c_{adv}$ in the advection-diffusion equation (upper row) and $\nu$ in the KdV-Burgers equation (lower row). Mean values are calculated over all test trajectories within each range, and error bars indicate $\pm 1$ standard deviation across those trajectories.*

## Conclusion

This work addressed the problem of coupled representation and evolution errors that arise when time-dependent PDEs are advanced on coarsened spatial grids. Our solution, the RECAST framework, combines a SHRED-delta component, which corrects the provisional coarse-grid solver update within the time-stepping loop, with the SHRED-SR component that reconstructs the corresponding fine-grid state from the corrected coarse-state history. Across six PDE systems with distinct combinations of transport, diffusion, dispersion, reaction, and wave dynamics, RECAST reduced the time-averaged relative error by approximately 50-92% compared with the uncorrected coarse-grid solver over 1000-step closed-loop rollouts, while recovering spatial structures that were not resolved by the coarse representation. The framework also maintained its improvement for unseen PDE parameter values in the advection–diffusion and KdV–Burgers cases as representative demonstrations. In comparison with an adapted $P^2C^2$Net baseline, RECAST achieved lower errors for both PDEs considered and maintained closer agreement with the fine-grid reference over an extended 5000-step rollout. These results show that correcting accumulated coarse-grid evolution error before reconstructing the fine-grid state can substantially extend the accuracy attainable from highly coarsened numerical simulations.

Although the present demonstrations are performed on small-scale one-dimensional PDEs, the observed $8\times$ to $16\times$ spatial coarsening has important implications for larger-scale simulations. In two dimensions, the same per-direction coarsening would reduce the number of grid nodes by $8^2$ to $16^2$, i.e., $64\times$to $256\times$; in three dimensions, the reduction would be $8^3$ to $16^3$, i.e., $512\times$to $4096\times$. Since conventional PDE solvers update the solution over the spatial degrees of freedom at every time step, such reductions could translate into substantial computational savings in higher-dimensional systems. The learned SHRED correction and super-resolution models instead operate on the reduced coarse state through global mappings applied once per time step, rather than sweeping a fine grid. As a result, while the practical speedup will depend on the underlying solver and inference costs, the scaling of the grid reduction indicates the potential computational benefit of extending RECAST to higher-dimensional problems.

Establishing this potential in practice will require extending the present proof-of-concept RECAST evaluation in several directions. First, the framework should be tested directly on two- and three-dimensional systems to determine whether the correction and reconstruction accuracy observed in the present paper is retained as the spatial state dimension and complexity increase. Evaluation on larger-scale problems and a broader range of PDEs, numerical solvers, and discretizations will also be needed to establish how generally RECAST transfers across different sources of coarse-grid error. Finally, the computational scaling should be assessed directly, including the wall-clock cost and inference overhead of the learned components relative to the savings obtained from coarse-grid evolution. These extensions, constituting the future work, will allow us to more fully determine the range of problems for which the accuracy improvements demonstrated by RECAST here translate into practical computational gains.

**Data and Code Availability**:

The codes for the RECAST framework and for generating the data used in this study are available on GitHub at: github.com/MariRe1992/recast.

## Appendix A: Long-horizon rollout comparison of RECAST and $P^2C^2$Net

This appendix extends the comparison with $P^2C^2$Net in Section 3.1 to a 5000-step rollout, providing a longer-horizon test of stability and accumulated trajectory error. The spatiotemporal field and profile comparisons in Figure 19 and Figure 20 show that RECAST continues to follow the fine-grid reference over the extended rollout, while the pure coarse-grid solver remains overly diffusive and phase-inaccurate. $P^2C^2$Net improves the early coarse prediction, but its trajectory progressively deteriorates at later times.

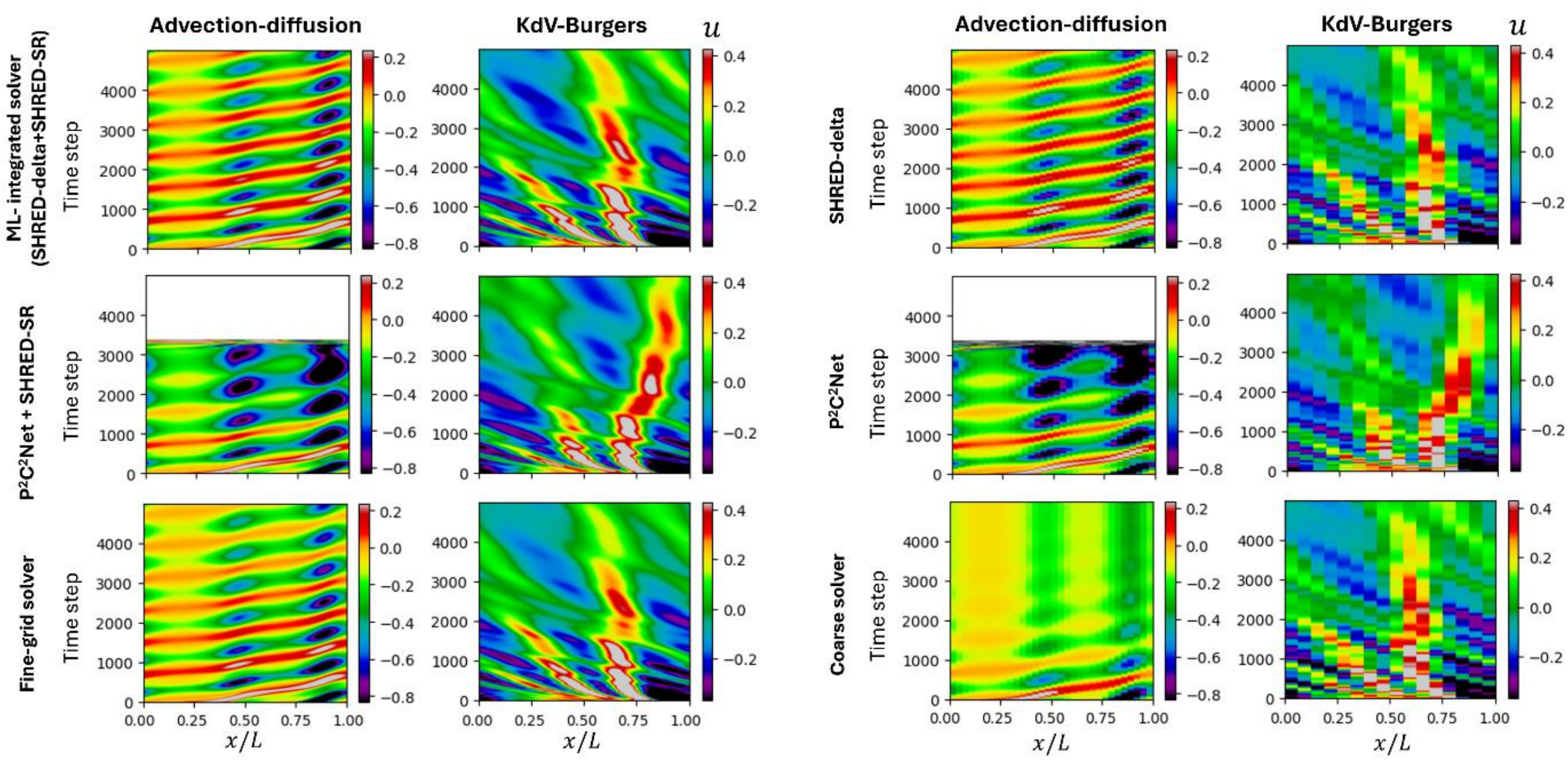

*Figure 19: Long-term rollout comparison of spatiotemporal solutions for advection-diffusion and KdV-Burgers test trajectories. The left group compares fine-grid fields obtained from RECAST, $P^2C^2$Net followed by offline SHRED-SR, and the fine-grid reference solver. The right group compares the corresponding coarse-grid trajectories from SHRED-delta, $P^2C^2$Net, and the pure coarse-grid solver.*

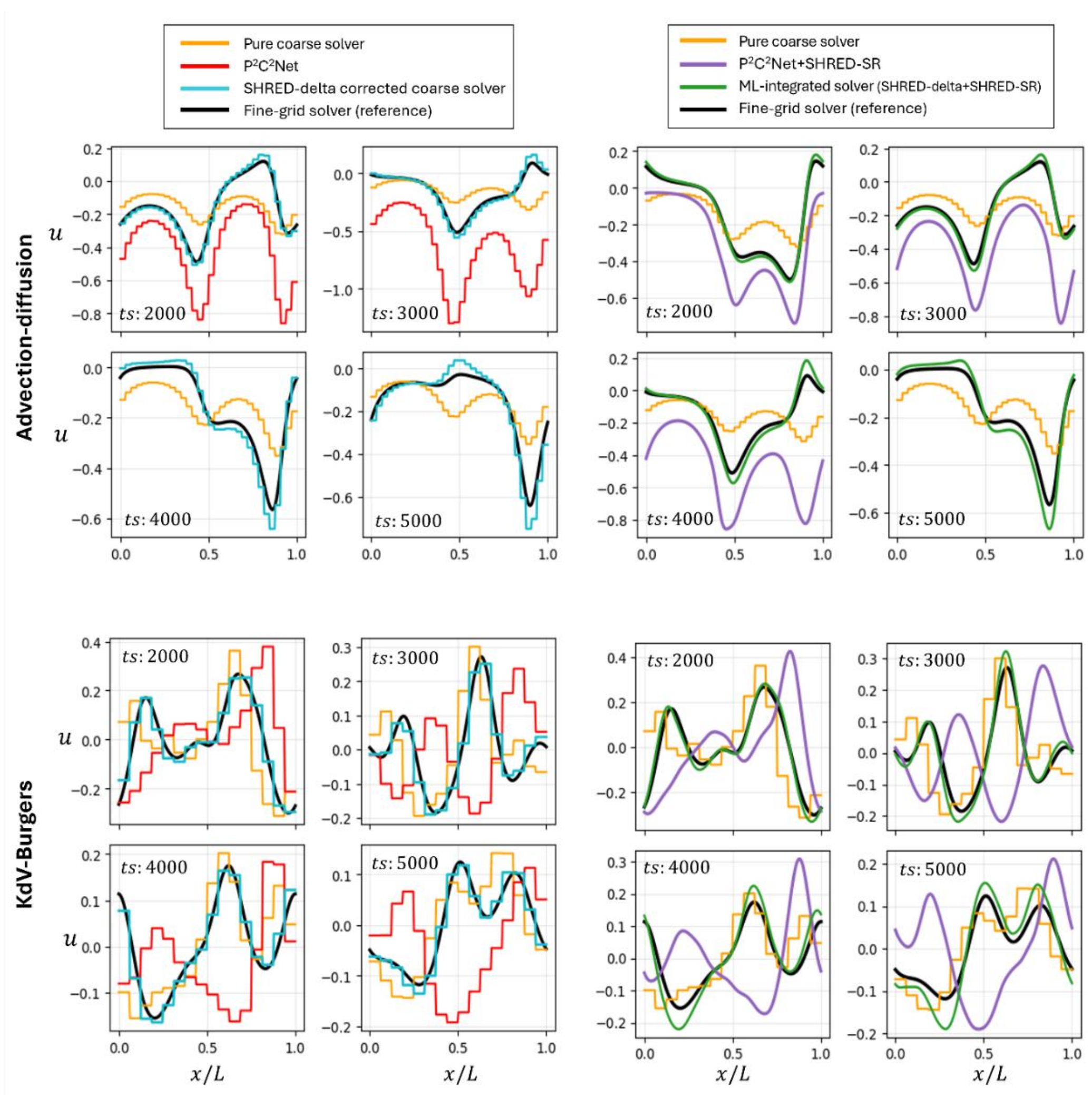

*Figure 20: Long-term rollout comparison of predicted spatial profiles at selected time steps for advection-diffusion and KdV-Burgers. Profiles are shown for the pure coarse-grid solver, $P^2C^2$Net, SHRED-delta corrected coarse solver, $P^2C^2$Net followed by offline SHRED-SR, RECAST, and the fine-grid reference solver.*

The long-term error traces in Figure 21 make the behavior observed in Figure 19 and Figure 20 clearer. For both PDEs, RECAST maintains substantially lower error growth than the baselines. $P^2C^2$Net accumulates error more rapidly and reaches or exceeds the pure coarse-solver error around, or shortly after, time step 1000 in both PDEs. In the advection–diffusion case, it subsequently becomes unstable after approximately 3000-3500 time steps.

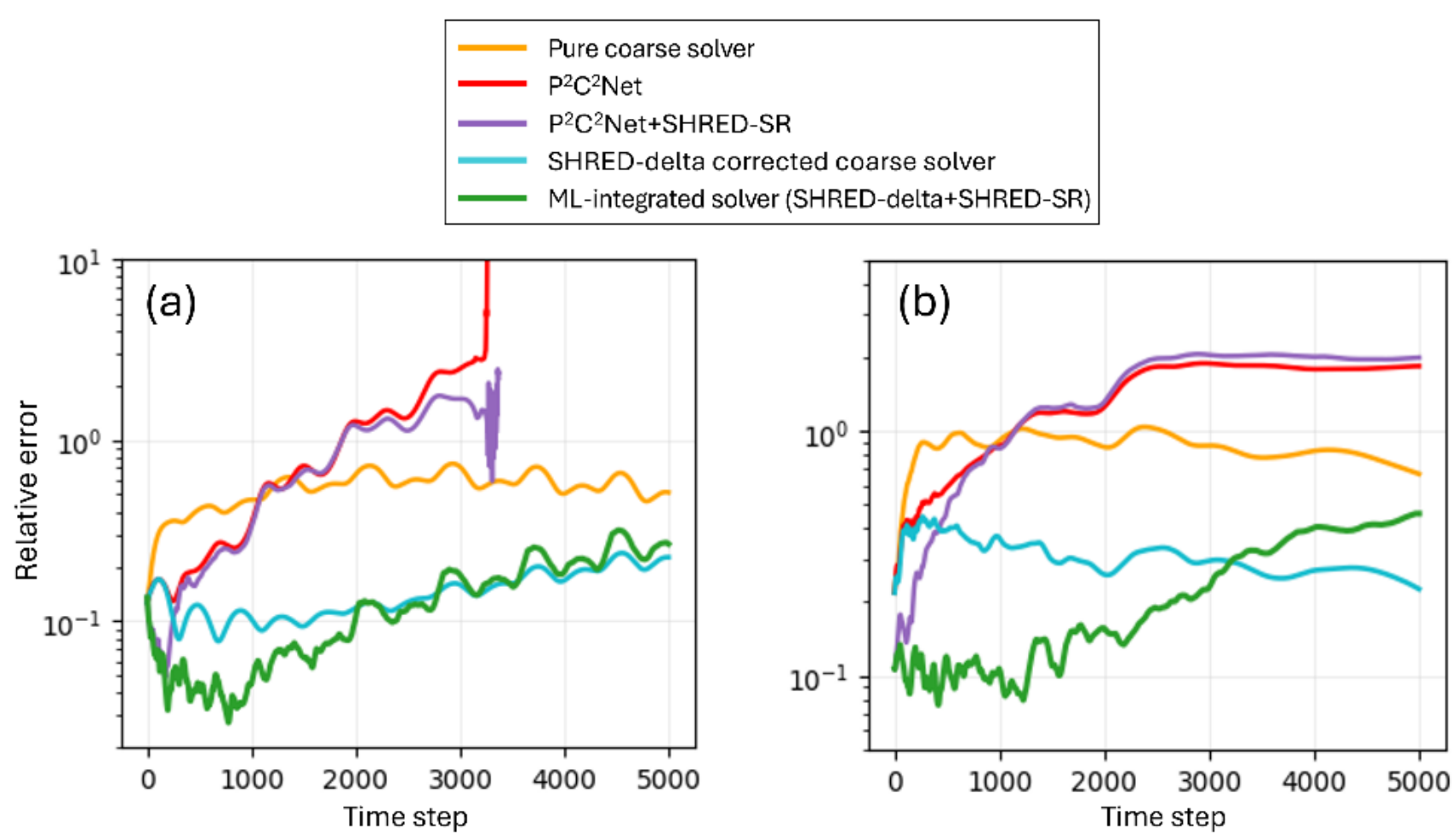

*Figure 21: Long-term relative error evolution (Eq. 30) for the presented test trajectories for (a) advection-diffusion and (b) KdV-Burgers. Errors are shown for the pure coarse-grid solver, $P^2C^2$Net, $P^2C^2$Net followed by offline SHRED-SR, the SHRED-delta corrected coarse solver, and RECAST.*

## Appendix B: Additional solution examples for unseen initial conditions

This appendix provides supplementary results for additional held-out advection–diffusion and KdV–Burgers test trajectories with unseen initial conditions sampled from the distributions in Table 2, extending the examples presented in Section 3.2.

The spatiotemporal solutions in Figure 22 and Figure 23 and spatial profiles in Figure 24 compare RECAST, the fine-grid reference solver, and the pure coarse-grid solver, providing further examples beyond those shown in the main text and demonstrating robustness across different initial-condition realizations.

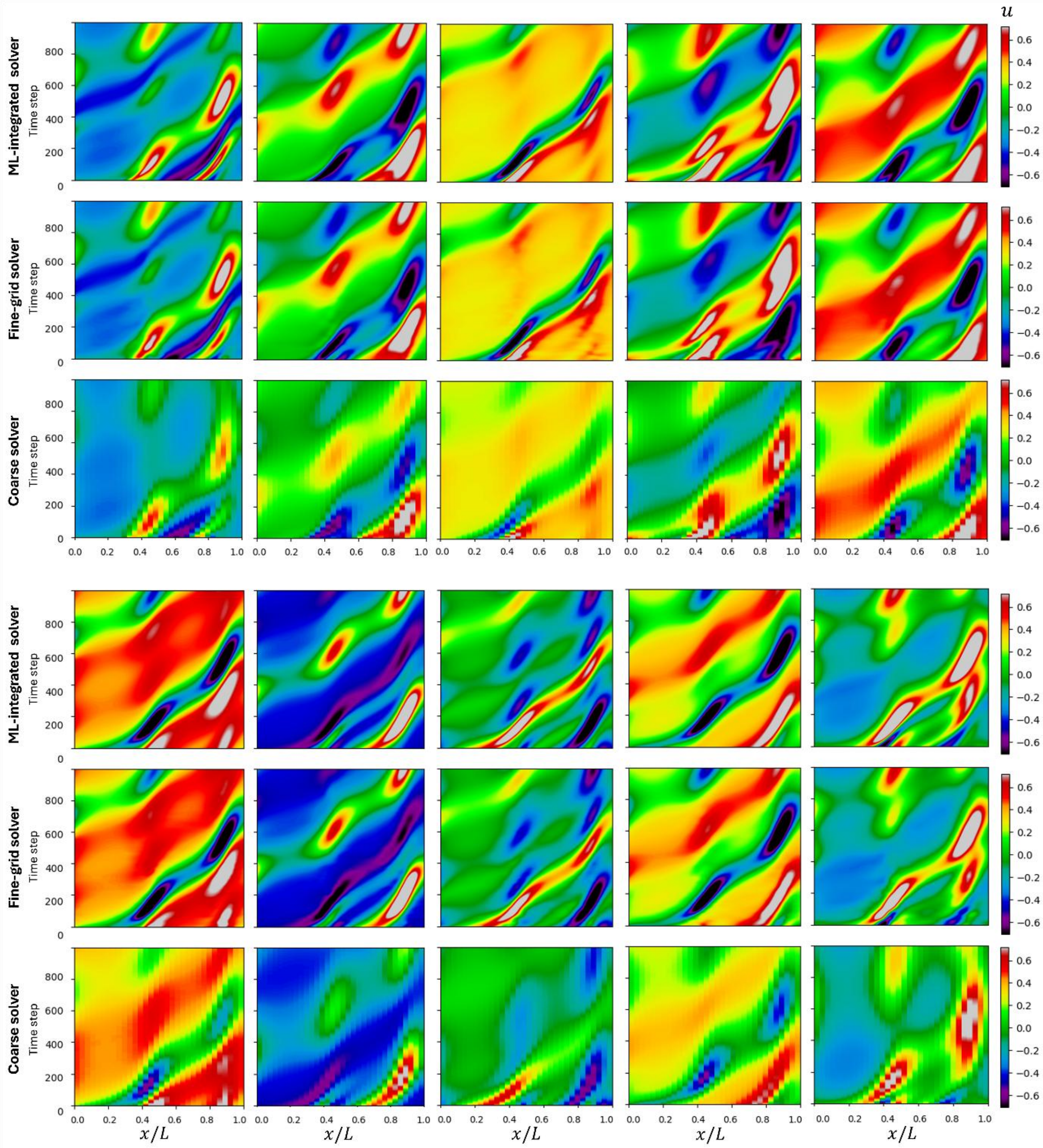

*Figure 22: Comparison of spatiotemporal solutions for different test initial conditions, obtained from RECAST, fine-grid reference solver, and pure coarse-grid solver for the advection-diffusion case.*

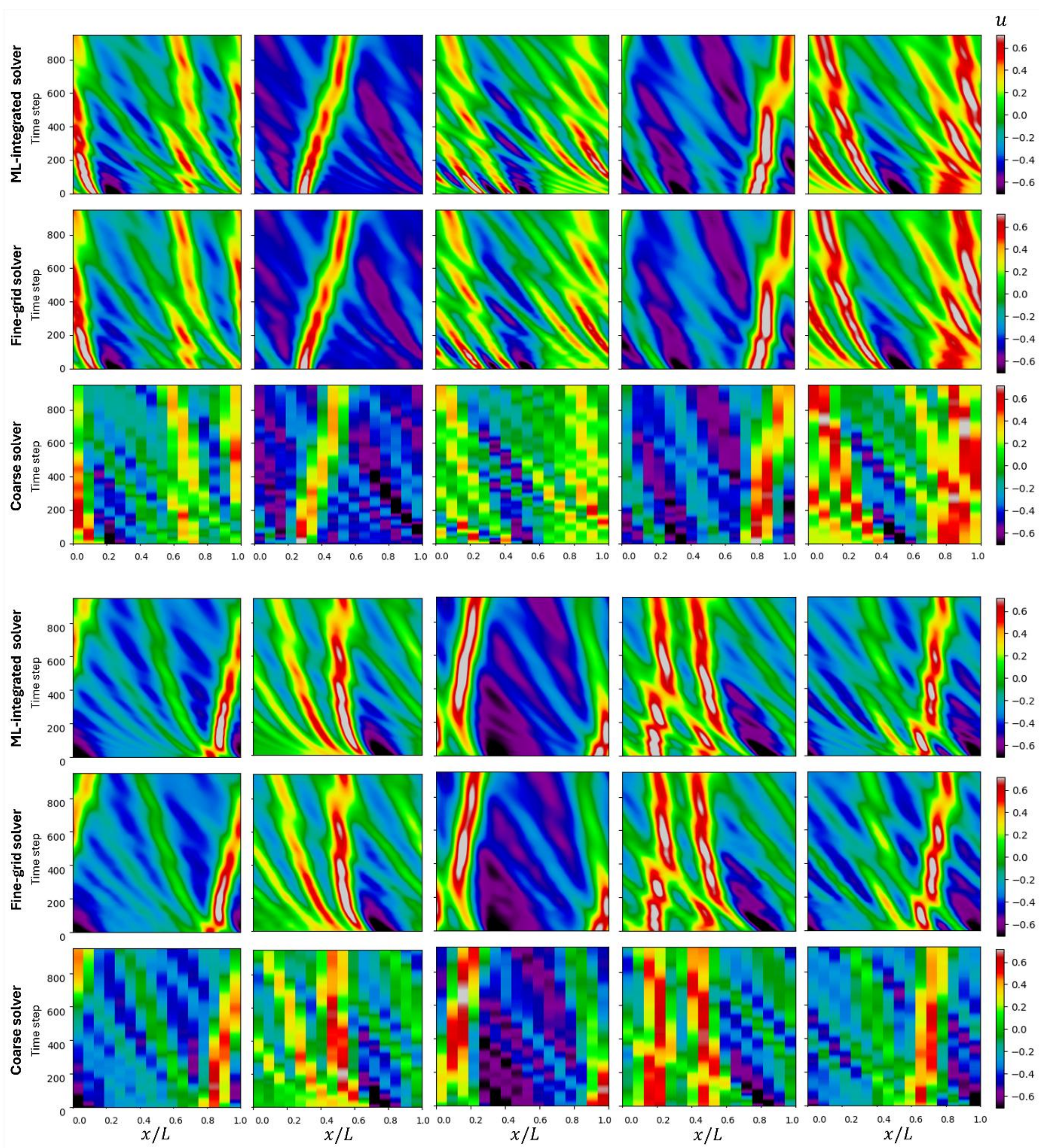

*Figure 23: Comparison of spatiotemporal solutions for different test initial conditions, obtained from RECAST, the fine-grid reference solver, and the pure coarse-grid solver for the KdV-Burgers case.*

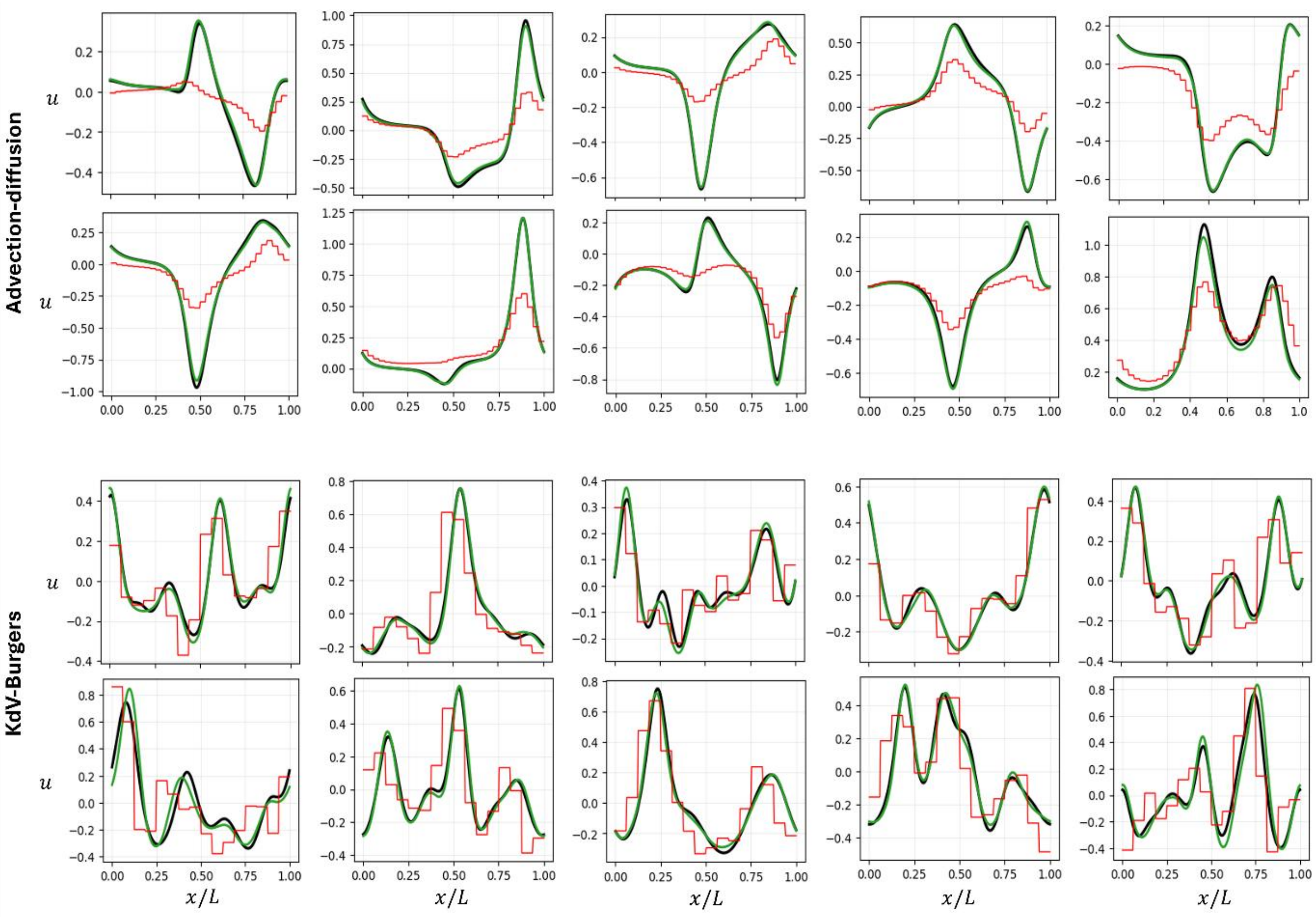

*Figure 24: Comparison of spatial profiles at time step 1000 for different test initial conditions, obtained from RECAST, the fine-grid reference solver, and the pure coarse-grid solver for advection-diffusion (upper rows) and KdV-Burgers (lower rows).*

## References:

[1] C. Dong, C. C. Loy, K. He, and X. Tang, "Image super-resolution using deep convolutional networks," *IEEE Transactions on Pattern Analysis and Machine Intelligence*, vol. 38, no. 2, pp. 295–307, 2016. DOI: 10.1109/TPAMI.2015.2439281.

[2] C. Ledig, L. Theis, F. Huszár, J. Caballero, A. Cunningham, A. Acosta, A. P. Aitken, A. Tejani, J. Totz, Z. Wang, and W. Shi, "Photo-realistic single image super-resolution using a generative adversarial network," in *2017 IEEE Conference on Computer Vision and Pattern Recognition (CVPR)*, Honolulu, HI, USA, pp. 105–114, 2017. DOI: 10.1109/CVPR.2017.19.

[3] X. Wang, K. Yu, S. Wu, J. Gu, Y. Liu, C. Dong, Y. Qiao, and C. C. Loy, "ESRGAN: Enhanced super-resolution generative adversarial networks," in *Computer Vision – ECCV 2018 Workshops*, Lecture Notes in Computer Science, vol. 11133, pp. 63–79, Springer, Cham, 2019. DOI: 10.1007/978-3-030-11021-5_5.

[4] K. Fukami, K. Fukagata, and K. Taira, "Super-resolution analysis via machine learning: A survey for fluid flows," *Theoretical and Computational Fluid Dynamics*, vol. 37, no. 4, pp. 421–444, 2023. DOI: 10.1007/s00162-023-00663-0

[5] F. Faraji and M. Reza, "Machine learning applications to computational plasma physics and reduced-order plasma modeling: A perspective," *Journal of Physics D: Applied Physics*, vol. 58, no. 10, article 102002, 2025. DOI: 10.1088/1361-6463/ada167.

[6] K. Fukami, K. Fukagata, and K. Taira, "Super-resolution reconstruction of turbulent flows with machine learning," *Journal of Fluid Mechanics*, vol. 870, pp. 106–120, 2019. DOI: 10.1017/jfm.2019.238.

[7] K. Fukami, K. Fukagata, and K. Taira, "Machine-learning-based spatio-temporal super resolution reconstruction of turbulent flows," Journal of Fluid Mechanics, vol. 909, article A9, 2021. DOI: 10.1017/jfm.2020.948.

[8] Z. Deng, C. He, Y. Liu, and K. C. Kim, "Super-resolution reconstruction of turbulent velocity fields using a generative adversarial network-based artificial intelligence framework," *Physics of Fluids*, vol. 31, no. 12, article 125111, 2019. DOI: 10.1063/1.5127031.

[9] M. Bode, M. Gauding, Z. Lian, D. Denker, M. Davidovic, K. Kleinheinz, J. Jitsev, and H. Pitsch, "Using physics-informed enhanced super-resolution generative adversarial networks for subfilter modeling in turbulent reactive flows," *Proceedings of the Combustion Institute*, vol. 38, no. 2, pp. 2617–2625, 2021. DOI: 10.1016/j.proci.2020.06.022.

[10] L. Lu, P. Jin, G. Pang, Z. Zhang, and G. E. Karniadakis, "Learning nonlinear operators via DeepONet based on the universal approximation theorem of operators," *Nature Machine Intelligence*, vol. 3, no. 3, pp. 218–229, 2021. DOI: 10.1038/s42256-021-00302-5.

[11] Z. Li, N. B. Kovachki, K. Azizzadenesheli, B. Liu, K. Bhattacharya, A. M. Stuart, and A. Anandkumar, "Fourier neural operator for parametric partial differential equations," in International Conference on Learning Representations (ICLR), 2021.

[12] M. Wei and X. Zhang, "Super-resolution neural operator," in *Proceedings of the IEEE/CVF Conference on Computer Vision and Pattern Recognition (CVPR)*, pp. 18247–18256, 2023. DOI: 10.1109/CVPR52729.2023.01750.

[13] J. Kossaifi, N. B. Kovachki, K. Azizzadenesheli, and A. Anandkumar, "Multi-grid tensorized Fourier neural operator for high-resolution PDEs," *Transactions on Machine Learning Research*, 2024.

[14] V. Oommen, S. Khodakarami, A. Bora, Z. Wang, and G. E. Karniadakis, "Learning turbulent flows with generative models for super resolution and sparse flow reconstruction," *Nature Communications*, vol. 17, no. 1, article 3707, 2026. DOI: 10.1038/s41467-026-70145-4.

[15] Y. Bar-Sinai, S. Hoyer, J. Hickey, and M. P. Brenner, "Learning data-driven discretizations for partial differential equations," *Proceedings of the National Academy of Sciences*, vol. 116, no. 31, pp. 15344–15349, 2019. DOI: 10.1073/pnas.1814058116.

[16] R. Maulik, O. San, A. Rasheed, and P. Vedula, "Subgrid modelling for two-dimensional turbulence using neural networks," *Journal of Fluid Mechanics*, vol. 858, pp. 122–144, 2019. DOI: 10.1017/jfm.2018.770.

[17] A. Beck, D. Flad, and C.-D. Munz, "Deep neural networks for data-driven LES closure models," *Journal of Computational Physics*, vol. 398, article 108910, 2019. DOI: 10.1016/j.jcp.2019.108910.

[18] D. Kochkov, J. A. Smith, A. Alieva, Q. Wang, M. P. Brenner, and S. Hoyer, "Machine learning–accelerated computational fluid dynamics," *Proceedings of the National Academy of Sciences*, vol. 118, no. 21, article e2101784118, 2021. DOI: 10.1073/pnas.2101784118.

[19] S. S. Blakseth, A. Rasheed, T. Kvamsdal, and O. San, "Deep neural network enabled corrective source term approach to hybrid analysis and modeling," *Neural Networks*, vol. 146, pp. 181–199, 2022. DOI: 10.1016/j.neunet.2021.11.021.

[20] K. Um, R. Brand, Y. Fei, P. Holl, and N. Thuerey, "Solver-in-the-loop: Learning from differentiable physics to interact with iterative PDE-solvers," *Advances in Neural Information Processing Systems*, vol. 33, pp. 6111–6122, 2020.

[21] B. List, L.-W. Chen, and N. Thuerey, "Learned turbulence modelling with differentiable fluid solvers: Physics-based loss functions and optimisation horizons," *Journal of Fluid Mechanics*, vol. 949, article A25, 2022. DOI: 10.1017/jfm.2022.738.

[22] T. Pfaff, M. Fortunato, A. Sanchez-Gonzalez, and P. W. Battaglia, "Learning mesh-based simulation with graph networks," in *International Conference on Learning Representations (ICLR)*, 2021.

[23] P. Ren, C. Rao, Y. Liu, J.-X. Wang, and H. Sun, "PhyCRNet: Physics-informed convolutional-recurrent network for solving spatiotemporal PDEs," *Computer Methods in Applied Mechanics and Engineering*, vol. 389, article 114399, 2022. DOI: 10.1016/j.cma.2021.114399.

[24] Q. Wang, P. Ren, H. Zhou, X.-Y. Liu, Z. Deng, Y. Zhang, R. Chengze, H. Liu, Z. Wang, J.-X. Wang, J.-R. Wen, H. Sun, and Y. Liu, "$P^2C^2$Net: PDE-preserved coarse correction network for efficient prediction of spatiotemporal dynamics," *Advances in Neural Information Processing Systems*, vol. 37, pp. 68897–68925, 2024. DOI: 10.52202/079017-2201.

[25] J. P. Williams, O. Zahn, and J. N. Kutz, "Sensing with shallow recurrent decoder networks," *Proceedings of the Royal Society A: Mathematical, Physical and Engineering Sciences*, vol. 480, no. 2298, article 20240054, 2024. DOI: 10.1098/rspa.2024.0054.

[26] M. Reza, F. Faraji, and J. N. Kutz, "Data-driven inference of high-dimensional spatiotemporal state of plasma systems," *Journal of Applied Physics*, vol. 136, no. 18, article 183301, 2024. DOI: 10.1063/5.0230056.

[27] F. Faraji, M. Reza, and J. N. Kutz, "Shallow recurrent decoder for reduced order modeling of E × B plasma dynamics," *Machine Learning: Science and Technology*, vol. 6, no. 2, article 025024, 2025. DOI: 10.1088/2632-2153/adcd20.

[28] M. Reza and F. Faraji, "Machine-learning-enabled full-state reconstruction of fusion plasmas from minimal sensor measurements," arXiv preprint arXiv:2607.04390, 2026. DOI: 10.48550/arXiv.2607.04390.

[29] M. Reza and F. Faraji, "Space-Time Information Interchangeability in Dynamical Systems: Conditions and Bounds for Replacing Spatial Sensors with Temporal Histories," arXiv preprint arXiv.2608.07728, 2026. DOI: 10.48550/arXiv.2608.07728.

[30] D. P. Kingma and J. Ba, "Adam: A method for stochastic optimization," in *3rd International Conference on Learning Representations (ICLR)*, San Diego, CA, USA, 2015.